\documentclass[10pt,twocolumn,letterpaper]{article}

\usepackage[pagenumbers]{cvpr} 

\usepackage{multirow}
\usepackage{algorithm}
\usepackage{algpseudocode}
\usepackage{rotating}
\usepackage{mathtools}
\usepackage{pifont}

\usepackage{soul}

\usepackage{color}
\usepackage{colortbl}
\usepackage{overpic}
\usepackage{pifont}
\definecolor{RedColor}{rgb}{0.949,0.275, 0.224}
\definecolor{OrangeColor}{rgb}{0.914,0.541,0.0.141}
\definecolor{GreenColor}{rgb}{0.137,0.573,0.565}

\newcommand{\shortname}{LASER}
\newcommand{\xmark}{\textcolor{RedColor}{\ding{55}}\xspace}
\newcommand{\cmark}{\textcolor{GreenColor}{\ding{51}}\xspace}

\newcommand\blfootnote[1]{%
  \begingroup
  \renewcommand\thefootnote{}\footnote{#1}%
  \addtocounter{footnote}{-1}%
  \endgroup
}

\newcommand{\inlinesection}[1]{\vspace{1mm} \noindent {\bf #1}}
\def\calW{\mathcal{W}}
\def\calS{\mathcal{S}}
\def\calG{\mathcal{G}}
\def\bC{\mathbf{C}}
\def\bI{\mathbf{I}}
\def\bP{\mathbf{P}}
\def\bT{\mathbf{T}}
\def\bR{\mathbf{R}}
\def\bt{\mathbf{t}}
\def\bp{\mathbf{p}}
\def\bq{\mathbf{q}}

\def\bu{\mathbf{u}}
\def\bv{\mathbf{v}}

\def\bCti{\bC_{t}^{(i)}}
\def\bPti{\bP_{t}^{(i)}}
\def\bCtiminusone{\bC_{t}^{(i-1)}}
\def\bPtiminusone{\bP_{t}^{(i-1)}}
\def\bTti{\bT_{t}^{(i)}}
\def\bRti{\bR_{t}^{(i)}}
\def\btti{\bt_{t}^{(i)}}
\def\siw{s_i^w}
\def\bRiw{\bR_i^w}
\def\btiw{\bt_i^w}

\def\barbDtiminusone{\bar{\mathbf{D}}_{t}^{(i-1)}}
\def\barbDti{\bar{\mathbf{D}}_{t}^{(i)}}
\def\barbPti{\bar{\mathbf{P}}_{t}^{(i)}}
\def\Mti{M_{t}^{(i)}}

\def\calO{\mathcal{O}}
\def\vx{\mathbf{x}}
\def\vy{\mathbf{y}}

\def\calC{\mathcal{C}}
\def\calL{\mathcal{L}}
\def\calE{\mathcal{E}}

\def\hatstni{\hat{s}_{t,n}^{(i)}}
\def\calCtni{\calC_{t,n}^{(i)}}
\def\calLtni{\calL_{t,n}^{(i)}}
\def\calLtmiminusone{\calL_{t,m}^{(i-1)}}

\definecolor{cvprblue}{rgb}{0.21,0.49,0.74}
\usepackage[pagebackref,breaklinks,colorlinks,allcolors=cvprblue]{hyperref}
\usepackage[normalem]{ulem}

\def\paperID{***} 
\def\confName{CVPR}
\def\confYear{2026}

\title{LASER: Layer-wise Scale Alignment for\\Training-Free Streaming 4D Reconstruction}

\author{%
  Tianye Ding$^{1*}$\quad
  Yiming Xie$^{1*}$\quad
  Yiqing Liang$^{2*}$\quad
  Moitreya Chatterjee$^3$ \quad
  Pedro Miraldo$^3$\quad
  Huaizu Jiang$^1$
  \\
  $^1$ Northeastern University \quad
  $^2$ Independent Researcher\quad
  $^3$ Mitsubishi Electric Research Laboratories\\
  \\
}

\begin{document}

\twocolumn[{%
\renewcommand\twocolumn[1][]{#1}%
\maketitle
\vspace{-10mm}
\includegraphics[width=1.0\linewidth]{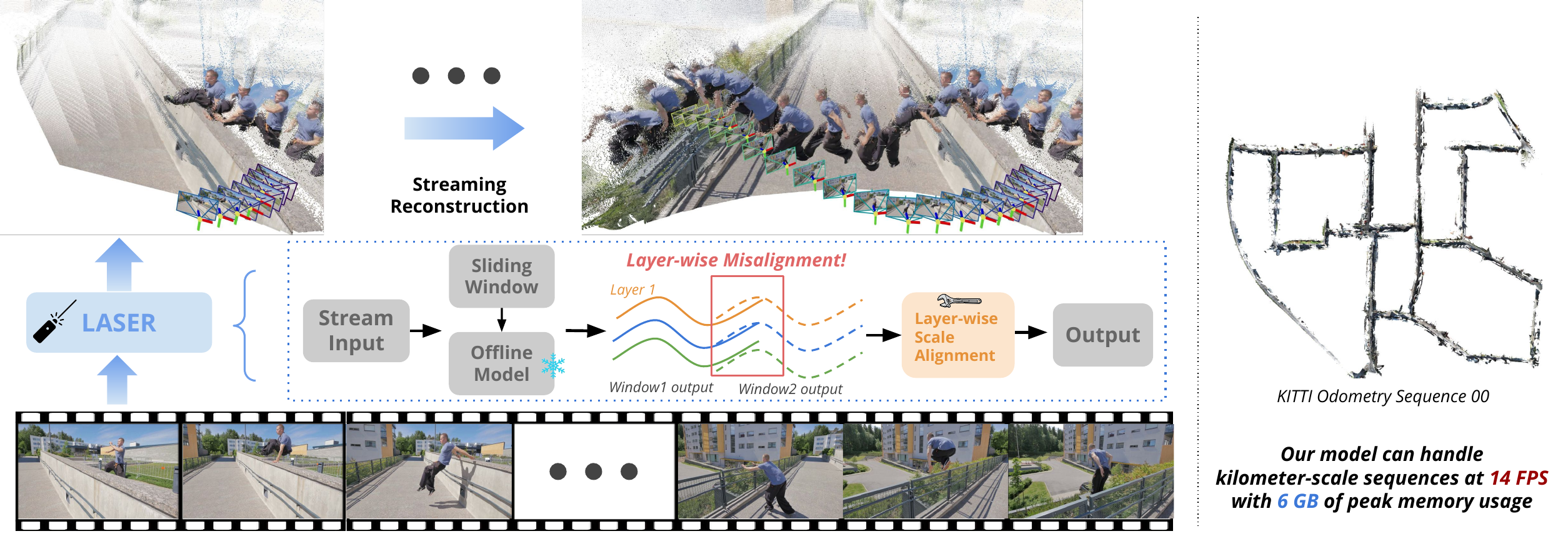}
    \captionof{figure}{
        {\bf \shortname} transforms offline reconstruction models into streaming systems via a sliding-window approach without retraining.
        Our submap registration and layer-wise scale alignment modules seamlessly align  windows into a globally consistent reconstruction.
}
    \vspace{5mm}
\label{fig:teaser}
}]

\blfootnote{$^*$ Equal contribution}
\vspace{-4mm}



\begin{abstract}
Recent feed-forward reconstruction models like VGGT and $\pi^3$ achieve impressive reconstruction 
quality but cannot process streaming videos due to quadratic memory complexity, limiting their practical deployment. 
While existing streaming methods address this through learned memory mechanisms 
or causal attention, they require extensive retraining and may not fully leverage 
the strong geometric priors of state-of-the-art offline models. We propose 
\shortname, a training-free framework that converts an offline reconstruction 
model into a streaming system by
aligning predictions across consecutive temporal windows. 
We observe that simple similarity transformation ($Sim(3)$) alignment 
fails due to layer depth misalignment: monocular scale ambiguity causes relative 
depth scales of different scene layers to vary inconsistently between windows. 
To address this, we introduce layer-wise scale alignment, which segments depth 
predictions into discrete layers, computes per-layer scale factors, and 
propagates them across both adjacent windows and timestamps.
Extensive experiments show 
that \shortname~achieves state-of-the-art performance on camera pose estimation 
and point map reconstruction 
while operating at 
14 FPS with 6 GB peak memory on a RTX A6000 GPU, enabling practical deployment for 
kilometer-scale streaming videos. 
Project website: \href{https://neu-vi.github.io/LASER/}{\texttt{https://neu-vi.github.io/LASER/}}
\end{abstract}    
\vspace{-7mm}
\section{Introduction}
\label{sec:intro}


Recovering 3D scene geometry from images has long been a central 
pursuit in computer vision, with applications ranging from robotic perception to 
digital cultural preservation. For decades, this problem was addressed through 
geometry-centric pipelines: 
Structure-from-Motion (SfM)~\cite{hartley2003multiple,schoenberger2016sfm} 
and Multi-View Stereo (MVS)~\cite{furukawa2009accurate,schoenberger2016mvs} systems with hand-crated designs.
While these classical methods achieve impressive 
accuracy with careful engineering, they remain sensitive to textureless regions and 
require known or estimated camera calibration. Recent advent of feed-forward 
neural approaches have fundamentally changed this landscape. DUSt3R~\cite{dust3r} 
pioneers direct regression of dense pointmaps from uncalibrated image pairs, 
eliminating the need for explicit correspondence solving. Subsequent work including 
VGGT~\cite{wang2025vggt} and $\pi^3$~\cite{wang2025pi} further handle arbitrary numbers of views, establishing a new paradigm where large-scale 
transformers trained on diverse data achieve superior reconstruction quality 
in zero-shot settings.


While these offline models achieve excellent reconstruction quality, 
they struggle with streaming scenarios due to quadratic memory 
complexity and the need to reprocess all frames when new observations 
arrive. Several recent works have proposed \emph{streaming variants} that 
process frames incrementally. Some approaches~\cite{wang2025cut3r,wang20243d,chen2025long3r} introduce persistent state or memory mechanisms for continuous 3D perception.
Another line of 
works~\cite{zhuo2025streaming,stream3r2025,li2025wint3r} adapt offline models with causal attention or combine sliding window with camera token pools.
Though effective, these methods 
share a common limitation: they require extensive retraining from scratch 
or through knowledge distillation to learn streaming-setting reconstruction, 
which is computationally expensive and may not fully leverage the strong 
geometric priors of state-of-the-art offline models like VGGT~\cite{wang2025vggt} 
and $\pi^3$~\cite{wang2025pi}. 
Moreover, recurrent designs like CUT3R~\cite{wang2025cut3r} can suffer from drift and catastrophic forgetting over long sequences~\cite{chen2025ttt3r}, while methods relying on growing memory face scalability constraints. 
Concurrent work VGGT-Long~\cite{deng2025vggtlongchunkitloop} also pursues a training-free approach by chunking sequences and aligning with $Sim(3)$, but as we show in \cref{sec:experiments}, simple rigid alignment is insufficient. 
Given that offline models already encode rich 3D priors, we ask: 
\emph{can we achieve both training-free conversion for streaming input and robust geometric alignment?}

In this work, we propose \textbf{\shortname}, a training-free framework that converts 
offline models into streaming systems by revisiting classical geometric 
principles. 
\shortname~ employs a sliding-window strategy, processing overlapping 
subsets of frames (windows) sequentially with a frozen offline model as the 
backbone.
Modern feed-forward models like VGGT~\cite{wang2025vggt} and 
$\pi^3$~\cite{wang2025pi} excel at producing accurate 3D reconstructions 
within individual windows. 
However, \emph{aligning these windows consistently} remains 
challenging. We observe that simple $Sim(3)$ alignment fails due to 
\emph{layer depth misalignment}: monocular scale ambiguity causes relative 
depth scales across scene layers (\eg, foreground \vs background) to vary 
between windows, particularly when camera translation is limited. A global 
$Sim(3)$ transformation applies uniform scaling to the entire window and cannot 
resolve such layer-wise scale variations. 
Drawing on classical insights that 
scenes naturally decompose into depth-ordered layers with distinct geometric 
properties~\cite{shade1998layered,wang1994representing,baker1998layered}, 
we propose \emph{layer-wise scale alignment} adapted to the modern deep learning reconstruction setting. 
Our approach segments reconstructed point maps into discrete layers using~\cite{felzenszwalb2004efficient}, computes per-layer scale factors between consecutive windows, and propagates these scales across the sequence to 
achieve layer-consistent alignment. 

Experimental results show that our design effectively addresses the 
aforementioned challenges and achieves state-of-the-art performance. 
\shortname~ 
outperforms 
the learned streaming methods while processing image streams at 14 FPS with only 6 GB peak memory on one RTX A6000 GPU. 
Notably, our training-free approach maintains competitive reconstruction quality with offline models (0.013$m$ vs 0.011$m$ mean accuracy on 7-Scenes~\cite{seven_scenes}) while enabling online processing.
This shows that when deep learning 
models provide strong local geometry, classical layer-based geometric reasoning 
can effectively unify their outputs into consistent long-range reconstructions 
without retraining.
Beyond quantitative gains, \shortname~offers significant practical advantages: it requires no model retraining, immediately applies to existing offline reconstruction models (as demonstrated with both VGGT~\cite{wang2025vggt} and $\pi^3$~\cite{wang2025pi} backbones), and scales to kilometer-long sequences that exceed the memory capacity of offline methods. 
As new, more powerful offline models emerge, \shortname~can immediately leverage their improvements without additional training costs, bridging the gap between the quality and efficiency of offline models and 
the streaming requirements.

Our main contributions are summarized as follows:
\begin{itemize}
\item We propose \shortname, a training-free framework for converting offline 
reconstruction models into streaming systems without retraining, 
(\eg, VGGT~\cite{wang2025vggt}, $\pi^3$~\cite{wang2025pi}).

\item We identify the layer depth misalignment problem arising from monocular 
scale ambiguity and propose layer-wise scale alignment to address it.

\item \shortname~achieves state-of-the-art performance in streaming pose estimation and 3D 
reconstruction ( $-68.6\%$ ATE on Sintel~\cite{sintel} pose estimation, $-63.9\%$ Acc on 7-Scenes~\cite{seven_scenes} reconstruction compared to the previous best) while operating at 14 FPS on an A6000 GPU with only 6 GB peak runtime memory. 
\end{itemize}

\section{Related Work}
\label{sec:related}



\begin{figure*}[th!]
    \centering
    \includegraphics[width=0.85\linewidth]{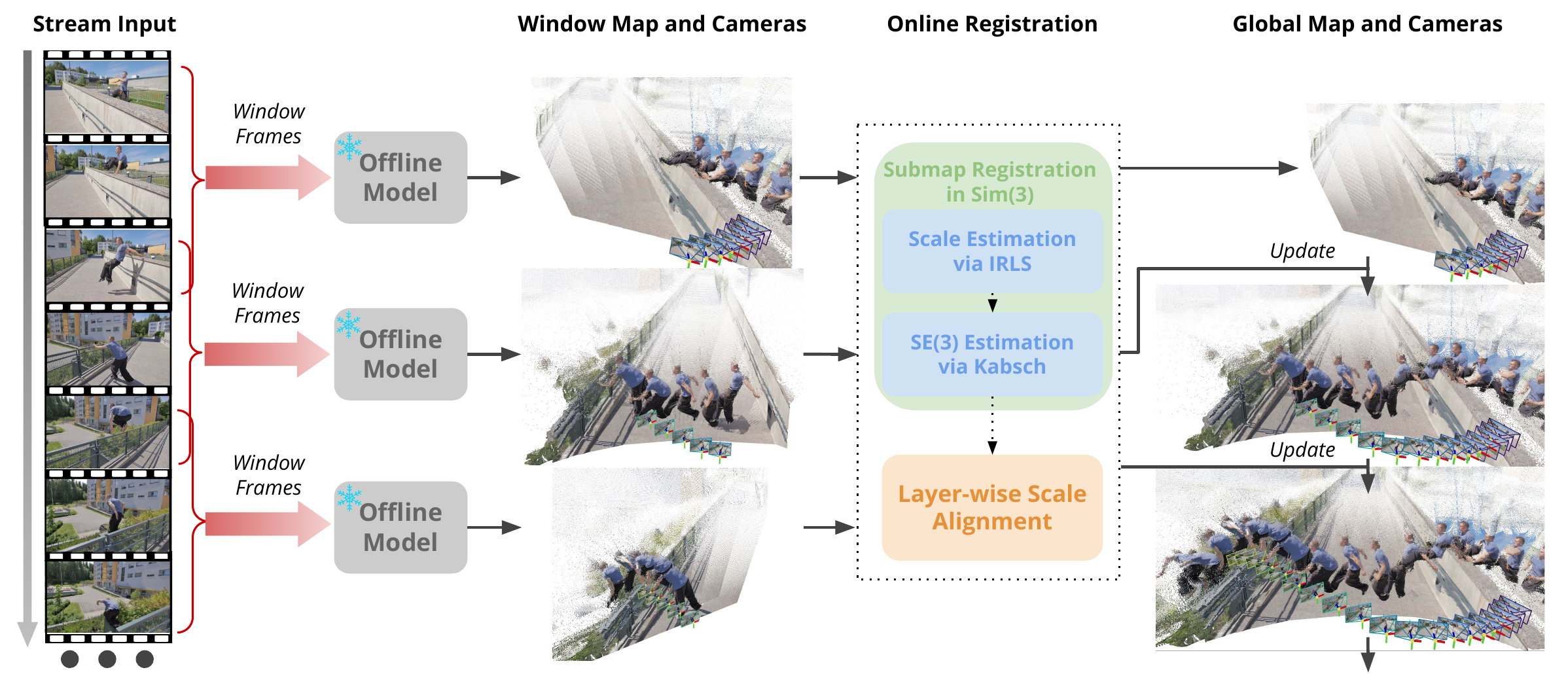}
    \caption{
        \textbf{Overview of \shortname.}
        \shortname~converts an offline reconstruction model to a streaming version without retraining.
        Given a video stream, we process frames in overlapping temporal windows with a frozen feed-forward reconstructor. 
        We incrementally register the submap to the global map with $\mathrm{Sim}(3)$ estimation and the proposed layer-wise scale alignment.
    }
    \vspace{-5mm}
\label{fig:pipeline}
\end{figure*}

\noindent\textbf{Learning-based 3D Reconstruction.}
Learning-based methods recast 3D reconstruction as a data-driven estimation problem rather than a purely geometric one.  
Early CNN-based pipelines~\cite{ji2017surfacenet,DeepMVS,yao2018mvsnet} replace handcrafted correspondence matching with learned feature aggregation and differentiable depth regression, paving the way toward end-to-end multi-view geometry learning.  
Subsequent 
methods~\cite{murez2020atlas,sun2021neucon} extend these ideas to online or large-scale reconstruction via recurrent fusion and volumetric integration.  
Implicit-field formulations~\cite{mildenhall2020nerf,wang2021neus,yu2022monosdf} achieve photorealistic surface and appearance modeling from sparse or monocular inputs, while explicit representations
\cite{kerbl3Dgaussians,triplane}
further improve rendering efficiency and scalability.
However, these models typically require per-scene optimization and are bounded by scene complexities~\cite{liang2025monocular}.
A parallel line of research seeks to eliminate 
the optimization 
through feed-forward geometric reasoning.  
DUSt3R~\cite{dust3r} pioneers a paradigm where 3D point clouds and relative poses are directly regressed from image pairs, and VGGT~\cite{wang2025vggt} generalizes this idea to variable-sized image sets with the help of learnable camera tokens.  
The recent $\pi^3$ model~\cite{wang2025pi} further introduces permutation-equivariant attention to unify structure and motion within a single scalable framework.  
Our proposed \shortname~builds on this feed-forward foundation, introducing a lightweight streaming formulation that adapts offline reconstruction models for continuous, efficient, kilometer-scale processing without retraining.

\noindent\textbf{Learning-based 4D Reconstruction.}
Learning-based 4D reconstruction approaches extend geometric and appearance modeling 
to temporal domain.
Early progress built on implicit neural representations~\cite{pumarola2020d,park2021nerfies,park2021hypernerf,li2020neural} introduces temporal conditioning or deformation fields.
To improve efficiency, subsequent approaches~\cite{WU_2024_CVPR, yang2023deformable3dgs, luiten2023dynamic, gaufre,kplanes_2023,Cao2023HexPlane,TiNeuVox,liu2024gear} apply similar extensions to explicit representations. 
Both implicit and explicit 4D representations require per-scene optimization 
, limiting usage in 
streaming settings.
Another series of works~\cite{han2025enhancing,zhang2025monstr,lu2024align3r,chen2025easi3r} extend feed-forward regression to dynamic settings.
\cite{st4rtrack2025,jin2024stereo4d,Liang2025ZeroShotMSF,li2024_MegaSaM} leverage dense video correspondence supervision to generalize to diverse dynamic scenes.
$\pi^3$~\cite{wang2025pi} unifies 3D and 4D reasoning through permutation-equivariant attention during large-scale training.
Our method shares the goal of scalable 
4D reconstruction
but focuses on the \emph{streaming} regime: adapting offline feed-forward geometry transformers for causal video processing. 
With a sliding-window formulation and 
geometry-aware 
alignment, \shortname~achieves temporally consistent reconstruction efficiently without retraining.

\begin{figure*}[t]
    \centering
    \includegraphics[width=.89\linewidth, trim={0 0 15pt 0}, clip]{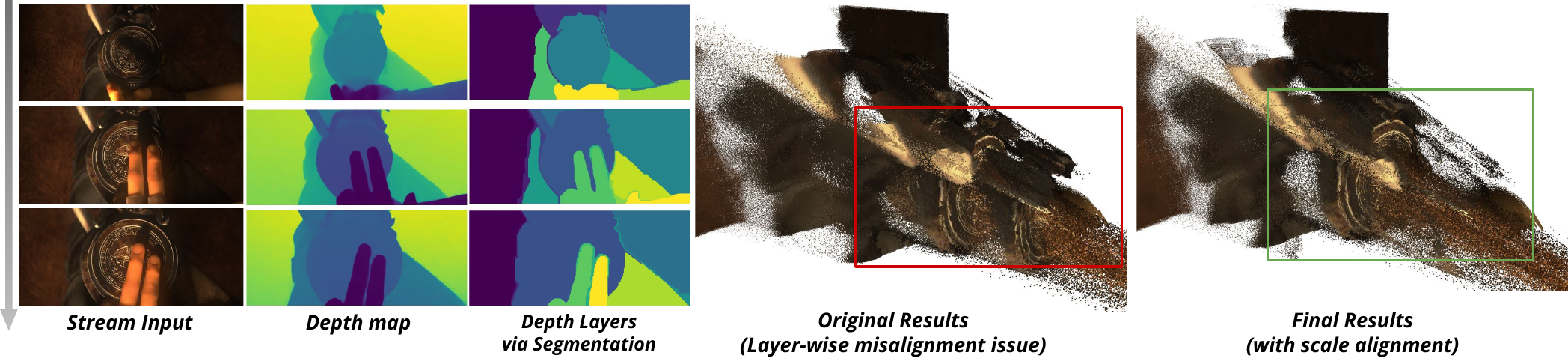}
    \caption{
        \textbf{Layer Depth Misalignment Issue}. 
        After the global $\mathrm{Sim}(3)$ alignment, surfaces at different depths may exhibit \textit{layer-wise scale inconsistency}: foreground regions appear over- or under-scaled relative to background structures across consecutive windows.
        This anisotropic scaling leads to visible distortions and metric drift in the fused reconstruction.
        We introduce Layer-wise Scale Alignment (LSA), a geometry-driven refinement that corrects distortions based on a layer graph.
    }
    \vspace{-4mm}
\label{fig:issue}
\end{figure*}

\noindent\textbf{Streaming Feed-Forward Reconstruction.}
Real-world applications such as autonomous driving, robotics, and AR/VR require models that process video streams efficiently and consistently. 
Recent works have tried to finetune offline feed-forward reconstructor to the streaming regime: 
memory-centric approaches introduce persistent or explicit spatial memory to extend temporal horizons~\cite{wang2025cut3r,wu_point3r_2025,spann3r}; 
causal-transformer designs process frames sequentially with token pooling or causal attention~\cite{li2025wint3r,stream3r2025,zhuo2025streaming}; 
test-time adaptation has also been explored for long videos~\cite{chen2025ttt3r}; 
parallel efforts bridge feed-forward prediction with SLAM-style optimization~\cite{mast3r_slam,vggtslam}. 
Aside from requiring retraining, these methods either accumulate scale drift over time, fail on long sequences due to memory limits or incur slow inference.
On the other hand, \cite{Yang_2025_Fast3R} pushes feed-forward reconstruction to the kilo-frame regime, but does not support streaming input.
\shortname\ addresses these limitations with a general, training-free framework that turns offline feed-forward models (\eg, VGGT or $\pi^3$) into a streaming system capable of handling kilo-frame dynamic sequences while preserving their reconstruction quality and efficiency.



\vspace{-2mm}
\section{Method}
\label{sec:method}
\subsection{Overview}
Our goal is to convert an offline 4D reconstruction model to a streaming version 
\emph{without retraining}.
Fig.~\ref{fig:pipeline} illustrates our pipeline. Given a video stream, we process frames in overlapping \emph{temporal windows}. Each window is passed through a frozen feed-forward reconstructor to predict {dense point maps} (local submaps) and {camera poses}.
We incrementally register the submap of the current window to the global map to complete the 
streaming 4D reconstruction.

\noindent\textbf{4D reconstruction in temporal windows.}
Let $\{\bI_t\}_{t=1}^{T}$ be a monocular RGB video with $T$ frames where each frame has a spatial dimension of $H\times W$. We form overlapping windows $\{\calW_i\}_{i=1}^{T/L}$, where each window contains $L$ consecutive frames. Let $a_i$ denote the start index of the frame of the $i$-th window; then
$\calW_i=\{t | a_i \leq t < a_i + L\},$
where $a_1 = 1$.
Two adjacent windows share an overlap of $O$ frames, \ie, $a_{i+1}=a_i+L-O$.


For each window $\calW_i$, a pretrained feed-forward reconstructor $f(\cdot)$ (\eg, VGGT~\cite{wang2025vggt}, $\pi^3$~\cite{wang2025pi}) predicts per-frame point maps and {camera poses}:
\begin{equation}
    f(\{\bI_t\}, \calW_i) = \{(\bPti, \bTti, \bCti) \mid a_i\leq t < a_i+L\},
\end{equation}
\vspace{-1mm}
where $\bPti \in \mathbb{R}^{H\times W\times 3}$ is a dense 3D point map in the window’s local coordinates and $\bTti=(\bRti | \btti)$ are the camera poses, consisting of a rotation matrix $\bRti\!\in\!SO(3)$ and a translation vector $\btti\!\in\!\mathbb{R}^3$, defined in the window's coordinate system. 
The reconstructor also outputs pixel-wise confidence scores $\bCti$,
which are used to form a set of mutually confident correspondences for scale estimation.
Based on $\bPti$ and $\bTti$, we construct the local submap $\calS_i$ 
for window $\mathcal{W}_i$ by transforming all per-frame point maps into the 
window's coordinate system. 


\vspace{0pt}
\noindent\textbf{Incremental global map reconstruction in the $\mathrm{Sim}(3)$ space.}
4D reconstruction in each window $\calW_i$ yields a local submap 
$\mathcal{S}_i=\{\bTti \bPti\}_{t\in \calW_i}$ 
in the window's own coordinate system.
We then estimate a similarity transform 
$(s_i^w,\bRiw,\btiw)\in\mathrm{Sim}(3)$ between $\calS_i$ to $\calG_{i-1}$, which are defined in the world coordinate system (in our case, the first temporal window's coordinate system, based on the estimated point maps of the overlapping region.
The induced camera pose in the world space for a frame $\bI_{t\in \calW_i}$ is
$\bT_{t}^{w} = (\bRiw \bRti |\ s_i^w \bRiw \btti + \btiw )$
The global map $\calG_i$ is then updated progressively as $\calG_i = \calG_{i-1} \cup \{\bT_{t}^{w} \bPti\}$, 
where $\calG_0=\emptyset$ in the initialization.

To estimate the $\mathrm{Sim}(3)$ transform, 
we first estimate the global scale factor $\siw$ via a robust IRLS (Iteratively Reweighted Least Squares) optimization~\cite{irls}, enforcing a shared metric across two adjacent windows.
Rotation and translation $(\bRiw,\btiw)$ are then optimized via the Kabsch algorithm ~\cite{Kabsch} under that metric using the \emph{scaled} camera anchors based on the estimated $\siw$.
We refer readers to the supplementary material for more details.


\subsection{Layer-wise Scale Alignment (LSA)}
\label{sec:lsa}

\vspace{-1mm}
Although the global $\mathrm{Sim}(3)$ registration aligns each window to a common scale,
it assumes isotropic scaling, where the same scale factor applies equally along all spatial axes.
In practice, this assumption breaks, \eg, under low-parallax motion, where a monocular reconstructor
cannot reliably constrain depth (the $Z$-axis) relative to lateral axes.
As a result, even after the global alignment, surfaces at different depths may exhibit
\emph{layer-wise scale inconsistency}: foreground regions appear over- or under-scaled
relative to background structures across windows, as shown in  Fig.~\ref{fig:issue}.
This anisotropic scaling along depth accumulates over time, leading to visible distortions
and metric drift in the fused reconstruction.
Following classical insights that scenes decompose into depth-ordered layers~\cite{shade1998layered,baker1998layered}, we introduce \emph{Layer-wise Scale Alignment (LSA)}, a geometry-driven refinement that corrects distortions based on a layer graph. 

\vspace{0pt}
\noindent\textbf{Depth layer extraction.} 
Inspired by classical layered representations, where a scene is decomposed into depth-ordered surfaces~\cite{shade1998layered,baker1998layered}, 
we extract depth layers by segmenting each depth map into spatially coherent 
regions at similar depths. 
Specifically, let $\barbPti=\bT_{t}^{w} \bPti\!\in\! \mathbb{R}^{H\times W\times 3}$ denote the 3D point map after the $\mathrm{Sim}(3)$ registration for frame $\bI_t$ in the temporal window $\calW_i$.
We derive a pseudo-depth map, denoted as $\barbDti$, from $\barbPti$ by taking its $Z$-coordinate components.
This pseudo-depth map is partitioned into $\Mti$ disjoint depth layers $\{\mathcal{L}_{t,m}^{(i)}\}_{m=1}^{\Mti}$ using an efficient segmentation algorithm~\cite{felzenszwalb2004efficient}. 
Each layer $\mathcal{L}_{t,m}^{(i)}$ corresponds to a continuous 
geometric surface patch with a coherent depth.
Examples are shown in Fig.~\ref{fig:issue}.

\begin{figure}[t!]
    \centering
\includegraphics[width=0.85\linewidth]{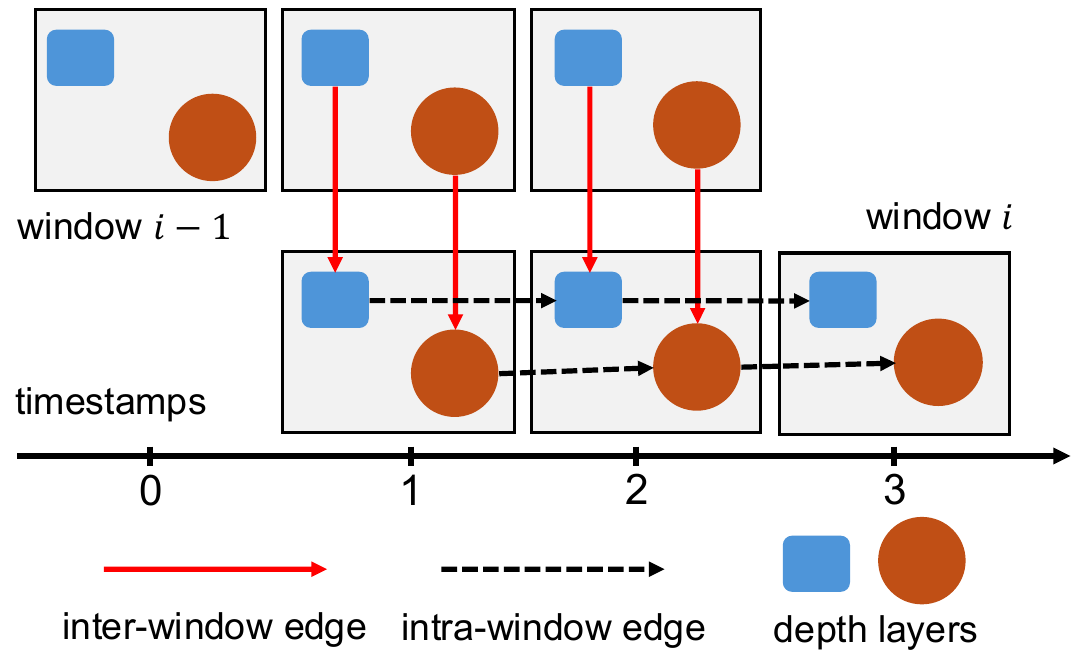}
    \vspace{-1mm}
    \caption{
        \textbf{Layer-wise Scale Alignment (LSA)}.
        We use a toy example to illustrate our proposed LSA.
    }
    \vspace{-5mm}
\label{fig:lsa}
\end{figure}

\vspace{0pt}
\noindent\textbf{Depth layer graph construction.}
\label{sec:layer-graph}
Let $\mathcal{O}_i \!=\! \calW_{i-1} \cap \calW_i$ be the set of overlapping timestamps.
To enforce consistent scaling between overlapping windows and across time,
we organize all depth layers into a \emph{directed graph} 
$\mathcal{H} = (\mathcal{V}, \mathcal{E})$, where the vertices correspond to the depth layers $\{\mathcal{L}_{t,m}^{(i-1)}\}_{t\in\calO_i}$ and $\{\mathcal{L}_{t,n}^{(i)}\}_{t\in\calW_i}$.
The edges $\mathcal{E}$ contains both inter-window and intra-window edges:
\begin{align*}
\mathcal{E}_{\mathrm{inter}} &\!=\!
\bigl\{(\calL_{t,m}^{(i-1)},\calL_{t,n}^{(i)})~\big|~
\mathrm{IoU}(\mathcal{L}_{t,m}^{(i-1)},\mathcal{L}_{t,n}^{(i)})>\tau, t\in\calO_i\bigr\},\\
\mathcal{E}_{\mathrm{intra}} &\!=\!
\bigl\{(\calL_{t-1,m}^{(i)},\calL_{t,n}^{(i)})~\big|~
\mathrm{IoU}(\mathcal{L}_{t-1,m}^{(i)},\mathcal{L}_{t,n}^{(i)})>\tau, t\in\calW_i\bigr\},
\end{align*}
with $\tau\!=\!0.3$.
$\mathcal{E}_{\mathrm{inter}}$ link layers at overlapping timestamps between windows $\mathcal{W}_{i-1}$ and $\mathcal{W}_i$, where the same geometric surface patch may appear under different scales due to independent per-window reconstruction.
Generally, the depth maps $\barbDtiminusone$ and $\barbDti$ for the same image $\bI_t$ in two windows are very close and the depth layers are almost identical.
Therefore, a single depth layer in $\calW_{i-1}$ will exactly be matched to another layer in $\calW_i$.
$\mathcal{E}_{\mathrm{intra}}$ connects the same depth layer across adjacent frames within window $\mathcal{W}_i$, encoding geometric continuity over time. 
An illustration is shown in Fig.~\ref{fig:lsa}.
Note that the edges are directed, pointing from a parent node to a child node across either the temporal windows or the timestamps. 
With this graph, we first estimate layer-wise scales based on $\mathcal{E}_{\mathrm{inter}}$ for the layers in the overlapping region.
The corrected scales will then be propagated and aggregated across both  $\mathcal{E}_{\mathrm{inter}}$ and $\mathcal{E}_{\mathrm{intra}}$.

\vspace{0pt}
\noindent\textbf{Layer-wise scale estimation via IRLS across inter-window edges.}
We estimate the layer-wise scales from point-wise correspondences in the intersection of two consecutive windows.
Specifically, we construct a set of correspondences $\calCtni=\{(d_p, d_q)\}$, where $d_p=\barbDtiminusone(x)$ and $d_q=\barbDti(x)$\footnote{We omit the symbol $x$ in $d_p$ and $d_q$ to avoid notation clutter.}, by using all depth values from pixel coordinate $x$ within the intersection of two layers $\calLtmiminusone \cap \calLtni$.
We then find the optimal scale for the layer $\calLtni$ by solving the following objective using IRLS.
\begin{equation}
\hatstni = \arg\min_{s>0}\;
\sum_{(d_p, d_q)\in\calCtni}
\rho~\!\bigl(\|\,s\,d_p - d_q\,\|\bigr),
\label{eq:layer_scale_irls}
\end{equation}
where $\rho(\cdot)$ is the Huber loss. 

\begin{algorithm}[t]
\caption{Layer-wise Scale Alignment}
\label{alg:lsa}
\begin{algorithmic}[1]
\Require Layer graph $\mathcal{H} = (\mathcal{V}, \mathcal{E})$ with
         vertices $\mathcal{V} = \{\calL_{t,m}^{(i-1)}\} \cup \{\calL_{t,n}^{(i)}\}$ and
         edges $\mathcal{E} = \mathcal{E}_{\mathrm{inter}} \cup \mathcal{E}_{\mathrm{intra}}$;
         temporal window $\calW_{i-1}$ and $\calW_i=\{t\}_{t=a_i}^{a_i + L}$.
\Ensure Final scales $\{s_{t,n}^{(i)}\}$ for layers $\{\calL_{t,n}^{(i)}\}$.

\State Initialize $A_{t,n}^{(i)} \gets 0$ and weight $W_{t,n}^{(i)} \gets 0$ for all $\calL_{t,n}^{(i)}$.

\Statex \textcolor{ForestGreen}{\textrm{\# inter-window scale optimization and propagation}}
\For{every $(\calL_{t,m}^{(i-1)}, \calL_{t,n}^{(i)}) \in \mathcal{E}_{\mathrm{inter}}$}
    \State compute $\hatstni$ according to Eq.(\ref{eq:layer_scale_irls})
    \State $w \gets \mathrm{IoU}(\calL_{t,m}^{(i-1)}, \calL_{t,n}^{(i)})$
    \State $A_{t,n}^{(i)} \gets A_{t,n}^{(i)} + w \cdot \hatstni$
    \State $W_{t,n}^{(i)} \gets W_{t,n}^{(i)} + w$
\EndFor

\Statex \textcolor{ForestGreen}{\textrm{\# temporal propagation along intra-window edges}}
\For{$t = a_i + 1$ \textbf{to} $a_i + L$}
    \For{every $(\calL_{t-1,m}^{(i)},\calL_{t,n}^{(i)}) \in \mathcal{E}_{\mathrm{intra}}$}
        \If{$W_{t-1,m}^{(i)} > 0$}
            \State $\mu_{t-1,m}^{(i)} \gets A_{t-1,m}^{(i)} / W_{t-1,m}^{(i)}$ \textcolor{ForestGreen}{\textrm{\# parent mean}}
            \State $w \gets \mathrm{IoU}(\calL_{t-1,m}^{(i)},\calL_{t,n}^{(i)})$
            \State $A_{t,n}^{(i)} \gets A_{t,n}^{(i)} + w \cdot \mu_{t-1,m}^{(i)}$
            \State $W_{t,n}^{(i)} \gets W_{t,n}^{(i)} + w$
        \EndIf
    \EndFor
\EndFor

\Statex \textcolor{ForestGreen}{\textrm{\# weighted average as the final scale for each layer}}
\For{every layer $\calL_{t,n}^{(i)}$}
    \State $s_{t,n}^{(i)} \gets A_{t,n}^{(i)} / W_{t,n}^{(i)}$ \textbf{if} $W_{t,n}^{(i)} > 0$ \textbf{else} 1
\EndFor

\end{algorithmic}
\end{algorithm}


\vspace{0pt}
\noindent\textbf{Scale propagation and aggregation along all the edges.}
After optimizing the layer-wise scales across inter-window edges $\calE_{\mathrm{inter}}$, for each layer $\calLtni$, the scales from its parent nodes along both $\calE_{\mathrm{inter}}$ and $\calE_{\mathrm{intra}}$ are propagated to it.
The layer $\calLtni$ may thus receive multiple scales, which will be aggregated in a weighted average manner. 
The weight for each edge is defined as the $\mathrm{IoU}$ score of two connected layers.
This procedure is summarized in Algorithm~\ref{alg:lsa}.
Please refer to the supplementary material for more elaboration.

Such layer-wise scale propagation and aggregation ensure the consistency across both adjacent windows and the temporal axis.
Once the layer-scale optimization is finished, each layer's scale will be propagated to its contained pixels, which will be used to adjust the reconstructed point map $\barbPti$.
As shown in Fig.~\ref{fig:issue}, it can effectively mitigate the distortions in the 4D reconstruction.

\vspace{-1mm}
\section{Experiments}
\label{sec:experiments}

We evaluate our proposed method, \shortname, against state-of-the-art approaches across three tasks: video depth estimation~(\cref{exp:video-depth}), camera pose estimation~(\cref{exp:camera-pose}), and multi-view point map estimation~(\cref{exp:point-map}) with details of the experimental setup in \cref{exp:setting}. 
Additional qualitative results are shown in \cref{sec:qualitative}. 
We further analyze model efficiency~(\cref{exp:efficiency}) and perform ablation studies to assess the contribution of each component~(\cref{exp:ablation}).


\subsection{Experimental Setup}
\label{exp:setting}

\subsubsection{Tasks, datasets, and metrics}
\paragraph{Video depth estimation protocol.}
Following \cite{zhang2025monstr, wang2025cut3r}, we evaluate on Sintel~\cite{sintel}, Bonn~\cite{bonn}, and KITTI~\cite{kitti}. 
Predicted depths are aligned to ground truth via \emph{scale-only} alignment.

\inlinesection{Camera pose estimation protocol.}
Following \cite{zhang2025monstr, wang2025cut3r}, we compare on small-scale Sintel~\cite{sintel}, ScanNet~\cite{scannet}, and TUM RGB-D~\cite{tum_rgbd}. 
Predicted trajectories are aligned to ground truth via a \emph{Sim(3)} transformation. 
For large-scale evaluation, we use KITTI Odometry~\cite{kitti} 
following
~\cite{deng2025vggtlongchunkitloop}.

\inlinesection{Multi-view point map estimation protocol.}
We run evaluation on 7-Scenes~\cite{seven_scenes} and NRGBD~\cite{nrgbd} with a keyframe sampling interval of 10, except for using an interval of 15 on NRGBD for \cite{zhuo2025streaming,stream3r2025} due to their memory limits. 
Predicted point maps are registered to ground truth using the Umeyama algorithm (coarse \emph{Sim(3)} alignment) followed by Iterative Closest Point (ICP) refinement. 

\begin{table}[t]
\centering
\setlength{\tabcolsep}{1.5pt}
\caption{\textbf{Video Depth Estimation on \textbf{Sintel}~\cite{sintel}, \textbf{Bonn}~\cite{bonn} and \textbf{KITTI}~\cite{kitti}.} We report Abs Rel and $\delta{<}1.25$.}
\vspace{-2mm}
\resizebox{\linewidth}{!}{
\begin{tabular}{lccc|cc|cc}
\toprule
& &  \multicolumn{2}{c|}{\textbf{Sintel}} & \multicolumn{2}{c|}{\textbf{Bonn}} & \multicolumn{2}{c}{\textbf{KITTI}}\\
\cmidrule(lr){3-4} \cmidrule(lr){5-6} \cmidrule(lr){7-8}
\textbf{Method} & Stream & Abs Rel $\downarrow$ & $\delta{<}1.25$ $\uparrow$ & Abs Rel $\downarrow$ & $\delta{<}1.25$ $\uparrow$ & Abs Rel $\downarrow$ & $\delta{<}1.25$ $\uparrow$ \\
\midrule 
VGGT~\cite{wang2025vggt}       &  \xmark    & 0.303 & \textbf{68.5} & 0.055 & 97.1 & 0.073 & 96.3 \\
$\pi^3$~\cite{wang2025pi}      &   \xmark     & \textbf{0.245} & 68.4 & \textbf{0.050} & \textbf{97.5} & \textbf{0.038} & \textbf{98.6}\\
\midrule
Spann3R~\cite{spann3r}      & \cmark  & 0.622 & 42.6 & 0.144 & 81.3 & 0.198 & 73.7  \\
CUT3R~\cite{wang2025cut3r}      & \cmark     & 0.421 & 47.9 & 0.078 & 93.7 & 0.118 & 88.1 \\
Point3R~\cite{wu_point3r_2025}    &  \cmark    & 0.452 & 48.9 & 0.060 & 96.0 & 0.136  & 84.2  \\
VGGT-SLAM~\cite{vggtslam} & \cmark& 0.424 & 56.0 & 0.076 & 93.2 & 0.136 & 81.8 \\
StreamVGGT~\cite{zhuo2025streaming}   & \cmark  & 0.323 & 65.7 & \underline{0.059} & \underline{97.2} & 0.173 & 72.1\\
STream3R$\beta$~\cite{stream3r2025} &\cmark & \underline{}{0.264} & \textbf{70.5} & 0.069 & 95.2 & \underline{0.080} & \underline{94.7} \\
WinT3R~\cite{li2025wint3r} &\cmark & 0.374 & 50.6 & 0.070 & 91.2 & 0.081 & 94.9 \\
TTT3R~\cite{chen2025ttt3r} &\cmark & 0.404 & 50.0 & 0.068 & 95.4 & 0.113 & 90.4 \\
\midrule
\textbf{VGGT+Ours}  &\cmark & 0.297 & 64.6 & 0.07 & 92.6 & 0.116 & 88.4 \\
\textbf{$\pi^3$+Ours} &\cmark & \textbf{0.247} & \underline{68.8} & \textbf{0.048} & \textbf{97.4} & \textbf{0.054} &  \textbf{98.3}\\
\bottomrule
\end{tabular}}
\vspace{-5mm}
\label{tab:scale_sintel_bonn}
\end{table}

\begin{table*}[t]
\centering
\setlength{\tabcolsep}{6pt}
\caption{\textbf{Camera Pose Estimation on \textbf{Sintel}~\cite{sintel}, \textbf{ScanNet}~\cite{scannet}, and \textbf{TUM}~\cite{tum_rgbd}.} We report ATE, translational RPE, and rotational RPE. }
\label{tab:traj_sintel_scannet_tum}
\vspace{-2mm}
\resizebox{.8\textwidth}{!}{
\begin{tabular}{lcccc ccc ccc}
\toprule
& & \multicolumn{3}{c}{\textbf{Sintel}} & \multicolumn{3}{c}{\textbf{ScanNet}} & \multicolumn{3}{c}{\textbf{TUM}} \\
\cmidrule(lr){3-5} \cmidrule(lr){6-8} \cmidrule(lr){9-11}
\textbf{Method} & Stream & ATE $\downarrow$ & RPE$_{\text{trans}}$ $\downarrow$ & RPE$_{\text{rot}}$ $\downarrow$
& ATE $\downarrow$ & RPE$_{\text{trans}}$ $\downarrow$ & RPE$_{\text{rot}}$ $\downarrow$
& ATE $\downarrow$ & RPE$_{\text{trans}}$ $\downarrow$ & RPE$_{\text{rot}}$ $\downarrow$ \\
\midrule
DUSt3R~\cite{dust3r}    & \xmark      & 0.290 & 0.132 & 7.869 & 0.246 & 0.108 & 8.210 & 0.140 & 0.106 & 3.286 \\
VGGT~\cite{wang2025vggt}    & \xmark        & \underline{0.171} & \underline{0.062} & \underline{0.471} & \underline{0.035} & \underline{0.015} & \underline{0.381} & \underline{0.012} & \underline{0.010} & \underline{0.309} \\
$\pi^3$~\cite{wang2025pi}     & \xmark        & \textbf{0.073} & \textbf{0.037} & \textbf{0.287} & \textbf{0.030} & \textbf{0.012} & \textbf{0.346} & \textbf{0.014} & \textbf{0.009} & \textbf{0.307} \\

\midrule
Spann3R~\cite{spann3r}   & \cmark      & 0.329 & 0.110 & 4.471 & 0.096 & 0.023 & 0.661 & 0.056 & 0.021 & 0.591 \\
CUT3R~\cite{wang2025cut3r}    & \cmark     & 0.213 & 0.066 & 0.621 & 0.099 & 0.022 & 0.600 & 0.046 & 0.015 & 0.473 \\
Point3R~\cite{wu_point3r_2025} & \cmark& 0.351 & 0.128 & 1.822 & 0.106 & 0.035 & 1.946 & 0.075 & 0.029 & 0.642 \\
VGGT-SLAM~\cite{vggtslam} & \cmark& 0.303 & 0.128 & 6.883 & 0.070 & 0.049 & 2.447 & 0.030 & 0.020 & 1.567 \\
StreamVGGT~\cite{zhuo2025streaming} & \cmark& 0.251 & 0.149 & 1.894 & 0.161 & 0.057 & 3.647 & 0.061 & 0.033 & 3.209  \\
STream3R$\beta$~\cite{stream3r2025} & \cmark& 0.213 & 0.076 & 0.868 & 0.052 & 0.021 & 0.850 & 0.026 & 0.013 & 0.330 \\
WinT3R~\cite{li2025wint3r} & \cmark& 0.225 & 0.097 & 1.092 & 0.062 & 0.020 & 0.690 & 0.074 & 0.023 & 0.774 \\
TTT3R~\cite{chen2025ttt3r} & \cmark& 0.201 & 0.063 & 0.617 & 0.064 & 0.021 & 0.592 & 0.028 & 0.012 & 0.379 \\
\textbf{VGGT+Ours}  & \cmark & \underline{0.131} & \underline{0.053}&	\underline{0.398}	& \underline{0.035}	&\underline{0.014}	&\underline{0.354}&	\textbf{0.013}	& \underline{0.010}	&\textbf{0.306} \\
\textbf{$\pi^3$+Ours} & \cmark & \textbf{0.061} & \textbf{0.028} & \textbf{0.249} & \textbf{0.031} & \textbf{0.012} & \textbf{0.339} & \underline{0.016} & \textbf{0.009} & \underline{0.308} \\
\bottomrule
\end{tabular}
}
\vspace{0mm}
\end{table*}
\begin{table*}[ht]
  \centering
  \vspace{-3mm}
  \caption{
  \textbf{Large-scale Camera Pose Estimation on \textbf{KITTI}~\cite{kitti}.} 
  We report ATE (lower is better). 
  CF: checkmark (\cmark) indicates no calibration required; 
  DR: checkmark (\cmark) indicates dense reconstruction supported. 
  We show metrics for sequence ID; \emph{Avg.} is the mean across sequences.
  Seq.~01 corresponds to a high-speed driving sequence whose motion differs from others; 
  \emph{Avg.$^*$} reports the mean ATE excluding Seq.~01. 
  \textit{OOM}: CUDA out-of-memory, \textit{TL}: tracking lost.
  }
  \vspace{-2mm}
  \resizebox{.9\textwidth}{!}{
  \begin{tabular}{l|cc|cc|ccccccccccc}
    \toprule
    \textbf{Method} & \textbf{CF} & \textbf{DR} & \textbf{Avg.} & \textbf{Avg.$^*$} & \textbf{00} & \textbf{01} & \textbf{02} & \textbf{03} & \textbf{04} & \textbf{05} & \textbf{06} & \textbf{07} & \textbf{08} & \textbf{09} & \textbf{10} \\
    \midrule

    ORB-SLAM2 (w/o LC)~\cite{murORB2}  & \xmark & \xmark & 69.73 & 26.48 & 40.65 & 502.20 & 47.82 & \textbf{0.94} & 1.30 & 29.95 & 40.82 & 16.04 & \underline{43.09} & \underline{38.77} & \textbf{5.42} \\
    ORB-SLAM2 (w/ LC)~\cite{murORB2}   & \xmark & \xmark & 54.82 & \textbf{9.46} & \textbf{6.03} & 508.34 & \textbf{14.76} & \underline{1.02} & 1.57 & \textbf{4.04} & \textbf{11.16} & \underline{2.19} & \textbf{38.85} & \textbf{8.39} & \underline{6.63} \\
    LDSO~\cite{gao2018ldso}             & \xmark & \xmark & \textbf{22.43} & \underline{23.50} & 9.32 & \underline{11.68} & \underline{31.98} & 2.85 & 1.22 & \underline{5.10} & 13.55 & 2.96 & 129.02 & 21.64 & 17.36 \\

    DROID-VO~\cite{teed2021droid}       & \xmark & \cmark & 54.19 & 51.19 & 98.43 & 84.20 & 108.80 & 2.58 & 0.93 & 59.27 & 64.40 & 24.20 & 64.55 & 71.80 & 16.91 \\
    DPVO~\cite{teed2023deep}          & \xmark & \xmark & 53.61 & 57.70 & 113.21 & 12.69 & 123.40 & 2.09 & \textbf{0.68} & 58.96 & 54.78 & 19.26 & 115.90 & 75.10 & 13.63 \\
    DROID-SLAM~\cite{teed2021droid}    & \xmark & \cmark & 100.28 & 75.85 & 92.10 & 344.60 & 107.61 & 2.38 & 1.00 & 118.50 & 62.47 & 21.78 & 161.60 & 72.32 & 118.70 \\
    DPV-SLAM~\cite{lipson2024deep}     & \xmark & \xmark & 53.03 & 57.19 & 112.80 & \textbf{11.50} & 123.53 & 2.50 & 0.81 & 57.80 & 54.86 & 18.77 & 110.49 & 76.66 & 13.65 \\
    DPV-SLAM++~\cite{lipson2024deep}  & \xmark & \xmark & \underline{25.75} & 27.14 & \underline{8.30} & 11.86 & 39.64 & 2.50 & \underline{0.78} & 5.74 & \underline{11.60} & \textbf{1.52} & 110.90 & 76.70 & 13.70 \\
    \midrule

    VGGT~\cite{wang2025vggt}            & \cmark & \cmark & / & / & \textit{OOM} & \textit{OOM} & \textit{OOM} & \textit{OOM} & \textit{OOM} & \textit{OOM} & \textit{OOM} & \textit{OOM} & \textit{OOM} & \textit{OOM} & \textit{OOM} \\
    $\pi^3$~\cite{wang2025pi}          & \cmark & \cmark & / & / & \textit{OOM} & \textit{OOM} & \textit{OOM} & \textit{OOM} & \textit{OOM} & \textit{OOM} & \textit{OOM} & \textit{OOM} & \textit{OOM} & \textit{OOM} & \textit{OOM} \\

    MASt3R-SLAM~\cite{mast3r_slam} & \cmark & \cmark & / & / & \textit{TL} & \textit{TL} & \textit{TL} & \textit{TL} & \textit{TL} & \textit{TL} & \textit{TL} & \textit{TL} & \textit{TL} & \textit{TL} & \textit{TL} \\
    CUT3R~\cite{wang2025cut3r} & \cmark & \cmark & / & / & \textit{OOM} & \textit{OOM} & \textit{OOM} & 148.07 & 22.31 & \textit{OOM} & \textit{OOM} & \textit{OOM} & \textit{OOM} & \textit{OOM} & \textit{OOM} \\
    Fast3R~\cite{Yang_2025_Fast3R}  & \cmark & \cmark & / & / & \textit{OOM} & \textit{OOM} & \textit{OOM} & \textit{OOM} & \textit{OOM} & \textit{OOM} & \textit{OOM} & \textit{OOM} & \textit{OOM} & \textit{OOM} & \textit{OOM} \\
    \midrule
    VGGT-Long~\cite{deng2025vggtlongchunkitloop} & \cmark & \cmark & \underline{27.64} & 18.28 & 8.67 & \textbf{121.17} & \textbf{32.08} & 6.12 & 4.23 & 8.31 & 5.34 & 4.63 & 53.10 & 41.99 & \underline{18.37} \\
    $\pi^3$-Long  & \cmark & \cmark & 30.72 & \underline{16.45} & \underline{6.28} & 173.45 & 63.92 & \underline{4.96} & \underline{1.66} & \underline{6.11} & \underline{4.89} & \underline{3.99} & \underline{36.08} & \textbf{14.78} & 21.84 \\
    \textbf{Ours ($\pi^3$)}  & \cmark & \cmark & \textbf{24.17} & \textbf{14.42} & \textbf{6.14} & \underline{121.61} & \underline{59.87} & \textbf{2.64} & \textbf{1.36} & \textbf{2.73} & \textbf{2.92} & \textbf{2.28} & \textbf{33.14} & \underline{17.95} & \textbf{15.20} \\
    \bottomrule
  \end{tabular}}
  \vspace{-4mm}
  \label{table:kitti_ate}
\end{table*}

\begin{figure*}[t]
\centering
\begin{minipage}{0.64\textwidth}
    \centering
    \setlength{\tabcolsep}{3pt}
    \captionof{table}{\textbf{Indoor, Short-term Multi-view Point Map Estimation on 7 scenes and NRGBD.} We report Accuracy (Acc, lower is better), Completeness (Comp, lower is better), and Normal Consistency (NC, higher is better)'s Mean and Median.}
    \label{tab:mvrecon_7scenes_nrgbd}
    \resizebox{\linewidth}{!}{%
    \begin{tabular}{l *{13}{c}}
    \toprule
    & & \multicolumn{6}{c}{\textbf{7 scenes}} & \multicolumn{6}{c}{\textbf{NRGBD}} \\
    \cmidrule(lr){3-8}\cmidrule(lr){9-14}
    & & \multicolumn{2}{c}{Acc$\downarrow$} & \multicolumn{2}{c}{Comp$\downarrow$} & \multicolumn{2}{c}{NC$\uparrow$}
    & \multicolumn{2}{c}{Acc$\downarrow$} & \multicolumn{2}{c}{Comp$\downarrow$} & \multicolumn{2}{c}{NC$\uparrow$} \\
    \cmidrule(lr){3-4}\cmidrule(lr){5-6}\cmidrule(lr){7-8}
    \cmidrule(lr){9-10}\cmidrule(lr){11-12}\cmidrule(lr){13-14}
    \textbf{Method} & Stream & Mean & Med. & Mean & Med. & Mean & Med. & Mean & Med. & Mean & Med. & Mean & Med. \\
    \midrule
    VGGT & \xmark & 0.017 & 0.005 & 0.024 & 0.01 & 0.586 & 0.633 & 0.013 & 0.006 & 0.011 & 0.003 & 0.705 & 0.837 \\
    Pi3 & \xmark & 0.011 & 0.004 & 0.019 & 0.008 & 0.598 & 0.652 & 0.012 & 0.005 & 0.01 & 0.003 & 0.704 & 0.826 \\
    \midrule
    CUT3R & \cmark & 0.036 & 0.019 & 0.029 & 0.008 & 0.624 & 0.695 & 0.135 & 0.069 & 0.055 & 0.012 & 0.684 & 0.825 \\
    StreamVGGT & \cmark & 0.043 & 0.022 & 0.028 & 0.008 & \textbf{0.639} & \textbf{0.719} & 0.082 & 0.051 & 0.046 & 0.013 & \underline{0.742} & \underline{0.918} \\
    STream3R$\beta$ & \cmark & 0.042 & 0.012 & 0.023 & 0.008 & \underline{0.631} & \underline{0.705} & 0.042 & 0.014 & 0.014 & 0.005 & \textbf{0.798} & \textbf{0.95} \\
    \textbf{VGGT+Ours} & \cmark & \underline{0.021} & \underline{0.007} & \underline{0.021} & \underline{0.007} & 0.590 & 0.639 & \textbf{0.020} & \underline{0.011} & \textbf{0.012} & \textbf{0.004} & 0.71 & 0.857 \\
    \textbf{$\pi^3$+Ours} & \cmark & \textbf{0.013} & \textbf{0.005} & \textbf{0.017} & \textbf{0.006} & 0.607 & 0.665 & \textbf{0.020} & \textbf{0.010} & \textbf{0.012} & \textbf{0.004} & 0.713 & 0.856 \\
    \bottomrule
    \end{tabular}%
    }
\end{minipage}%
\hfill
\begin{minipage}{0.34\textwidth}
    \centering
    \includegraphics[width=\linewidth]{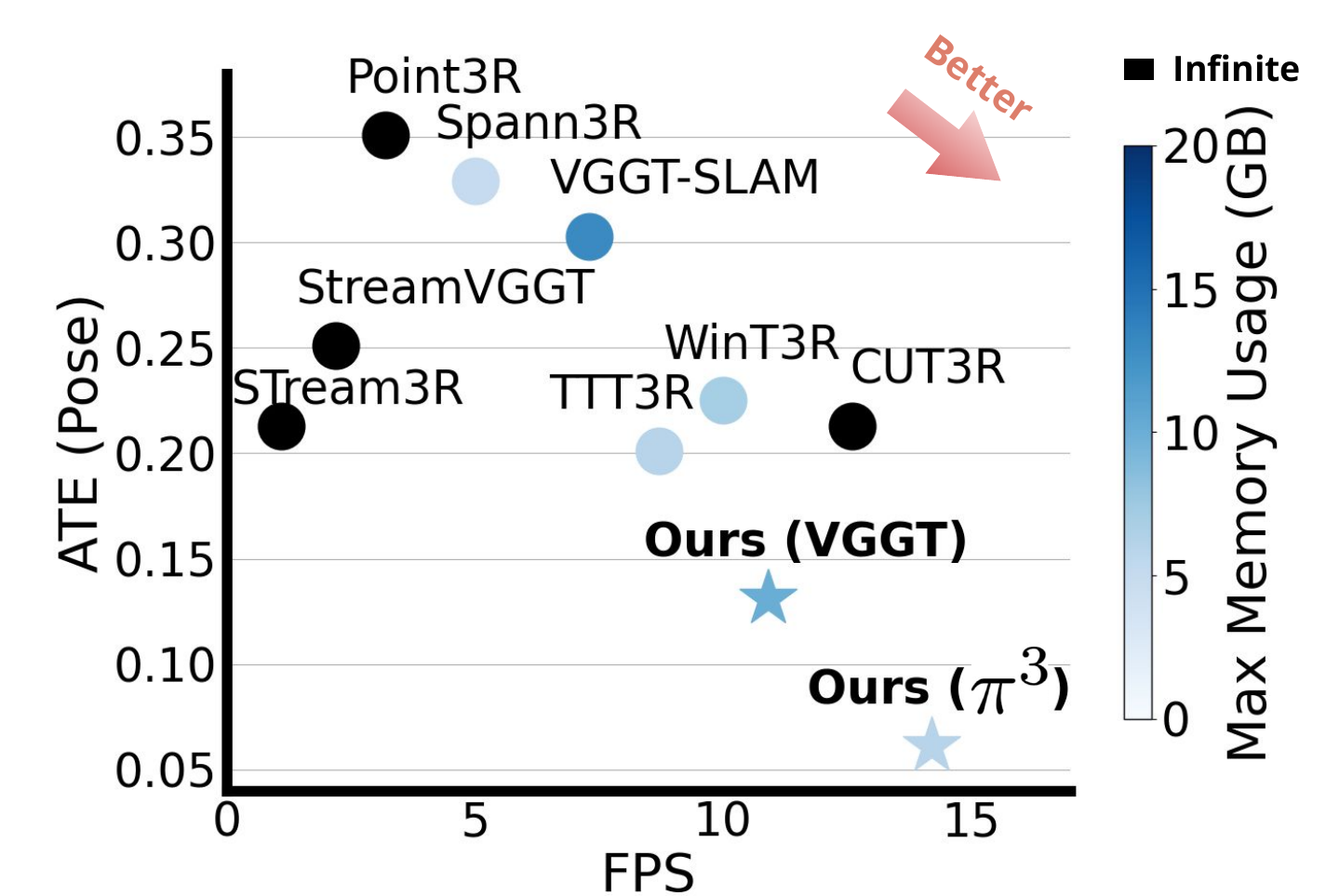}
    \vspace{-6mm}
    \captionof{figure}{We report the running FPS (on a RTX A6000 GPU), peak memory usage, and pose estimation error (ATE).}
    \label{fig:efficiency}
\end{minipage}
\end{figure*}

\begin{table}[t]
\centering
\setlength{\tabcolsep}{6pt}
\vspace{-1mm}
\caption{\textbf{Ablation Studies for Layer-wise Scale Alignment~(\cref{sec:lsa}).} 
\emph{wo$/$ LSA} denotes not using LSA component.
\emph{w$/$ SAM 2} denotes replacing the efficient segmentation algorithm~\cite{felzenszwalb2004efficient} with a recent counterpart SAM~2.
\emph{wo$/$ $\mathcal{E}_{\mathrm{intra}}$} denotes removing the temporal propagation step from LSA. 
}
\vspace{-2mm}
\resizebox{.4\textwidth}{!}{
\begin{tabular}{lcc|cc}
\toprule
& \multicolumn{2}{c|}{\textbf{Sintel}} & \multicolumn{2}{c}{\textbf{Bonn}}  \\
\cmidrule(lr){2-3} \cmidrule(lr){4-5}
 & Abs Rel $\downarrow$ & $\delta{<}1.25$ $\uparrow$ & Abs Rel $\downarrow$ & $\delta{<}1.25$ $\uparrow$ \\
\midrule 
\textbf{Ours} & \textbf{0.247} & \textbf{68.8} & \textbf{0.048} & \textbf{97.4} \\
wo$/$ LSA & 0.328 & 51.4 & 0.123 & 85.6 \\
w$/$ SAM 2~\cite{ravi2024sam2} & 0.251 & 67.8 & 0.051 & 97.4 \\
wo$/$ $\mathcal{E}_{\mathrm{intra}}$ & 0.261 & 64.7 & 0.05 & 97.1 \\
\bottomrule
\end{tabular}}
\vspace{-1mm}
\label{tab:ablation_lsa}
\end{table}
\begin{table}[t]
\centering
\setlength{\tabcolsep}{6pt}
\vspace{-2mm}
\caption{\textbf{Ablation on LSA~(\cref{sec:layer-graph})'s IoU threshold $\tau$.}}
\vspace{-2mm}
\resizebox{.35\textwidth}{!}{
\begin{tabular}{lcc|cc}
\toprule
& \multicolumn{2}{c|}{\textbf{Sintel}} & \multicolumn{2}{c}{\textbf{Bonn}} \\
\cmidrule(lr){2-3} \cmidrule(lr){4-5}
\textbf{$\tau$} & Abs Rel $\downarrow$ & $\delta{<}1.25$ $\uparrow$ & Abs Rel $\downarrow$ & $\delta{<}1.25$ $\uparrow$ \\
\midrule 
0.2 & 0.249 & 68.7 & 0.051 & 94.7 \\
\textbf{0.3 (default)} & 0.247 & 68.8 & 0.048 & 97.4 \\
0.4 & 0.247 & 68.4 & 0.048 & 97.4 \\
0.5 & 0.249 & 68.0 & 0.048 & 97.4 \\
0.6 & 0.248 & 68.0 & 0.044 & 97.1 \\
\bottomrule
\end{tabular}}
\vspace{-6mm}
\label{tab:ablation_iou}
\end{table}

\vspace{-1mm}
\subsubsection{Baselines}
\vspace{-1mm}
\inlinesection{Offline feed-forward models.} We include DUSt3R~\cite{dust3r}, Fast3R~\cite{Yang_2025_Fast3R}, VGGT~\cite{wang2025vggt}, and $\pi^3$~\cite{wang2025pi}, which process static image batches without temporal constraints. 
\inlinesection{Streaming or online feed-forward methods.}
We include Spann3R~\cite{spann3r}, CUT3R~\cite{wang2025cut3r}, MASt3R-SLAM~\cite{mast3r_slam}, Point3R~\cite{wu_point3r_2025}, VGGT-SLAM~\cite{vggtslam}, StreamVGGT~\cite{zhuo2025streaming}, STream3R$\beta$~\cite{stream3r2025}, WinT3R~\cite{li2025wint3r}, and TTT3R~\cite{chen2025ttt3r}, which enable causal inference or maintain persistent memory.

\inlinesection{Classical SLAM systems.}
For camera pose evaluation on KITTI~Odometry, we compare to SLAM methods including ORB-SLAM2~\cite{murORB2}, LDSO~\cite{gao2018ldso}, DROID-VO, DROID-SLAM~\cite{teed2021droid}, DPV-SLAM, and DPV-SLAM++~\cite{lipson2024deep}. 

\inlinesection{Training-free concurrent work.}
To further demonstrate the strength of our training-free design and ensure that performance is not dominated by a strong backbone, we also evaluate against VGGT-Long~\cite{deng2025vggtlongchunkitloop}, a concurrent training-free streaming framework built on VGGT~\cite{wang2025vggt}. 
For a fair comparison, we re-implement its pipeline on the $\pi^3$~\cite{wang2025pi} backbone, denoted as $\pi^3$-Long, enabling a one-to-one comparison under identical base models.

\vspace{-1mm}
\subsubsection{Implementation Details}

We instantiate \shortname\ using either VGGT~\cite{wang2025vggt} or $\pi^3$~\cite{wang2025pi} as the offline 4D reconstruction backbone. 
On the kilometer-scale KITTI~Odometry, we incorporate loop closure following the VGGT-Long~\cite{deng2025vggtlongchunkitloop} configuration for fair comparison.
More details are in the supplemental material.

\vspace{-1mm}

\subsection{Video Depth Estimation}
\label{exp:video-depth}
\cref{tab:scale_sintel_bonn} shows the video depth estimation results.
For streaming methods, a slight degradation on performance is reasonable compared with offline methods, since streaming methods operate with limited temporal context which leads to inconsistent depth predictions.

Compared to prior streaming baselines such as CUT3R~\cite{wang2025cut3r}, StreamVGGT~\cite{zhuo2025streaming}, and STream3R$\beta$~\cite{stream3r2025}, \shortname\ achieves the lowest Abs~Rel across all three datasets when compared to the best-performing baseline on each,
as well as the highest $\delta{<}1.25$ accuracy on Bonn 
and KITTI 
, while ranking second on Sintel.
Across all datasets, \shortname\ maintains the performance of its offline backbones VGGT~\cite{wang2025vggt} and $\pi^3$~\cite{wang2025pi} while operating in the streaming setting. 
These results demonstrate that \shortname\ delivers high-fidelity depth estimation across diverse domains while operating fully in a streaming manner.
Note that many baseline methods, such as StreamVGGT, Stream3R$\beta$, and VGGT-SLAM, also build upon offline approaches, \emph{yet their performance degrades significantly compared to the offline counterparts.}


\vspace{-1mm}
\subsection{Camera Pose Estimation}
\label{exp:camera-pose}
\cref{tab:traj_sintel_scannet_tum} reports results on small-scale datasets. 
On all three datasets, \shortname~($\pi^3$)
achieves the best results in almost all metrics and even surpasses its offline backbones in several cases. 
This demonstrates the effectiveness of our framework in pose estimation.
\shortname~(VGGT) consistently ranks second across all metrics, further validating the generality and robustness of our framework across backbones.

On large-scale outdoor sequences 
(\cref{table:kitti_ate}), \shortname\ using $\pi^3$ as backbone achieves the second-lowest mean ATE among all methods on both Avg.\ and Avg.$^*$ metrics. 
It attains accuracy comparable to or better than well-designed SLAM systems such as ORB-SLAM2~\cite{murORB2} and DROID-SLAM~\cite{teed2021droid}, which yield only sparse reconstructions and may require camera calibration. 
Meanwhile, dense offline models like VGGT~\cite{wang2025vggt} and $\pi^3$~\cite{wang2025pi} fail to process long sequences due to memory limits, and streaming variants such as CUT3R~\cite{wang2025cut3r} and MASt3R-SLAM~\cite{mast3r_slam} either run out of memory or lose tracking. In contrast, \shortname\ remains stable across all eleven sequences, producing globally consistent trajectories.
\shortname~also outperforms training-free streaming concurrent work VGGT-Long~\cite{deng2025vggtlongchunkitloop} and its variant $\pi^3$-Long by a $12$–$21\%$ reduction in avg ATE. 
Notably, although $\pi^3$-Long shares the same backbone with ours, it performs worse, indicating that the improvement is from our streaming algorithm design rather than backbone capacity. 



\subsection{Multi-View Point Map Estimation}
\label{exp:point-map}

\cref{tab:mvrecon_7scenes_nrgbd} reports short-term multi-view point map estimation results. 
\shortname\ consistently improves Acc and Comp over prior streaming baselines. 
While NC of $\pi^3$+Ours is slightly lower than that of StreamVGGT or STream3R$\beta$, 
this difference arises from the $\pi^3$ backbone’s limited surface-normal fidelity. 
Nevertheless, our formulation, despite training-free, also improves NC over the $\pi^3$.
A similar pattern is observed between VGGT+Ours and VGGT.
These results indicate that our online integration produces smoother, more coherent surface orientations than the backbone. 

\vspace{-1mm}

\subsection{Qualitative Results}
\label{sec:qualitative}
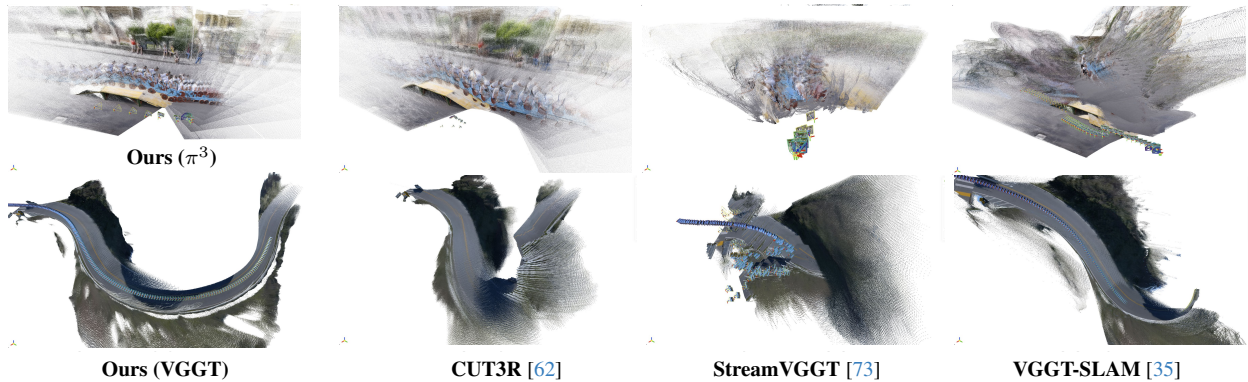
\begin{figure*}[t!]
\footnotesize
    \centering
    \vspace{-1mm}
    \begin{overpic}[trim={0 40pt 0 670pt},clip, width=0.95\linewidth]{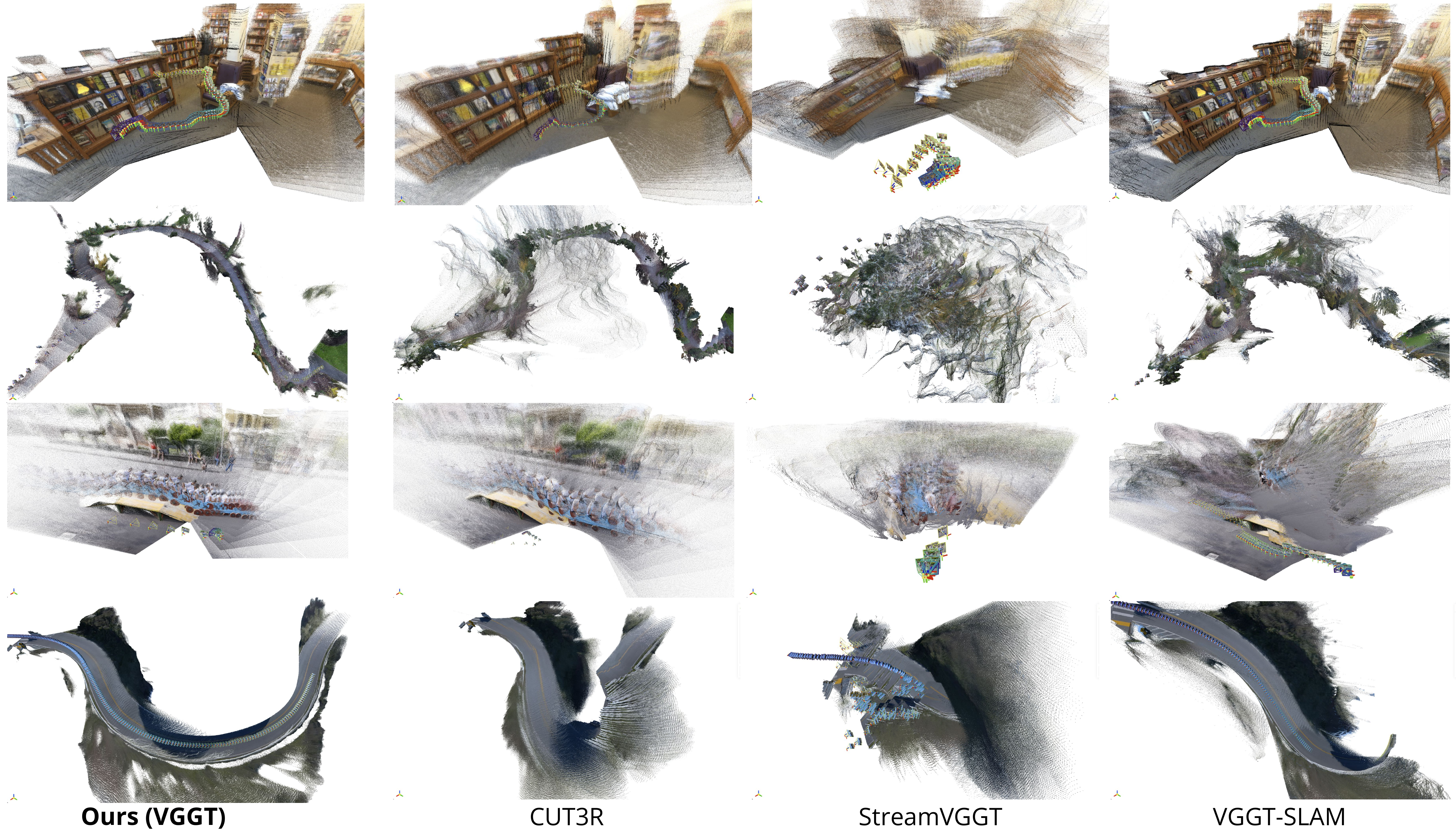}
    \put(10,15){\textbf{Ours~($\pi^3$)}}
    \put(8,-2){\textbf{Ours~(VGGT)}}
    
    \put(36,-2){\textbf{CUT3R}~\cite{wang2025cut3r}}
    \put(57,-2){\textbf{StreamVGGT}~\cite{zhuo2025streaming}}
    \put(81,-2){\textbf{VGGT-SLAM}~\cite{vggtslam}}
    \end{overpic}
    
    \vspace{2mm}
    \caption{
        Qualitative comparisons on 
        \textbf{DAVIS}~\cite{davis} and \textbf{Hike}~\cite{meuleman2023localrf} (top to bottom).  
        For each sequence, we show the reconstructed global point cloud with the estimated camera 
        trajectory overlaid. \emph{Please zoom in for camera trajectory details}.  
    }
    \vspace{-5mm}
\label{fig:vis}
\end{figure*}
We present qualitative comparisons in \cref{fig:vis}. 
Across all methods, our approach produces noticeably sharper scene geometry and more accurate camera trajectories. 
The examples 
cover
conditions of fast-motion videos and large-scale outdoor environments, highlighting robustness under diverse viewpoints and motions. 
These results demonstrate that \shortname\ generalizes well across datasets, delivering dense and stable reconstructions without any retraining or per-scene optimization.





\subsection{Efficiency Analysis}
\label{exp:efficiency}
\shortname~also demonstrates strong efficiency, as shown in \cref{fig:efficiency}. All experiments are performed on a RTX A6000 GPU. Compared to streaming feed-forward baselines, our method achieves the highest runtime speed ($\sim$14.2 FPS) with only 6 GB of peak memory usage when using $\pi^3$~\cite{wang2025pi} as our offline model, while maintaining superior performance in video depth and camera pose estimation.
When using VGGT~\cite{wang2025vggt}, we also have a competitive inference speed of $\sim$10.9 FPS and 10 GB peak memory usage.

\subsection{Ablation Studies}
\label{exp:ablation}

We evaluate key components of our pipeline through a series of ablations, using $\pi^3$~\cite{wang2025pi} as the backbone.


\inlinesection{Layer-wise Scale Alignment (\cref{sec:lsa}).}
\cref{tab:ablation_lsa} investigates the components of the LSA module on video depth estimation, as LSA does not affect camera pose estimation.  
Disabling LSA leads to clear drops in depth accuracy.  
We also try substituting the segmentation algorithm~\cite{felzenszwalb2004efficient} with SAM~2~\cite{ravi2024sam2}. Despite trading speed for better segmentation, SAM~2 does not improve accuracy.  
Finally, disabling propagation through $\mathcal{E}_{\text{intra}}$ ignores temporal relationships and prevents scale updates in non-overlapping frames, which harms global consistency across long sequences.

\begin{figure}[t]
    \centering
\includegraphics[trim={0 35pt 0 20pt}, clip, width=1.0\linewidth]{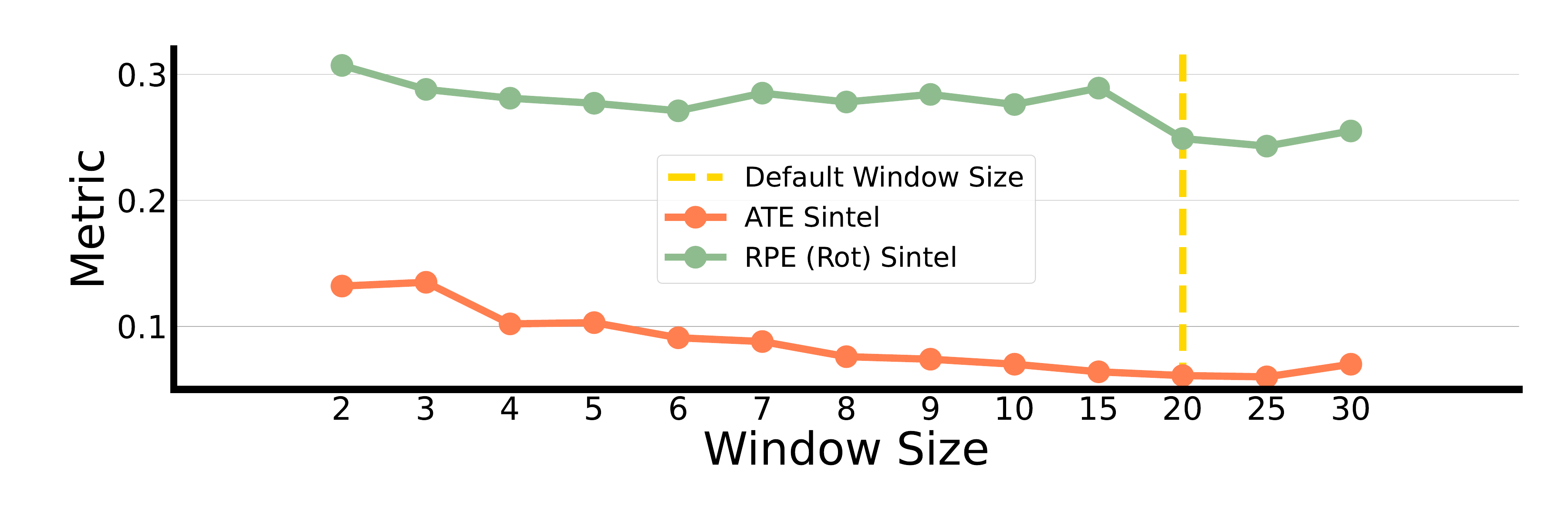}
    \vspace{-8mm}
    \caption{
        \textbf{Ablation on window size $L$.} In default, we use $L=20$.
    }
    \vspace{-5mm}
\label{fig:ablation_window}
\end{figure}

\inlinesection{Hyperparameters.}
\cref{fig:ablation_window}
and \cref{tab:ablation_iou} examine the effect of key hyperparameters: window size $L$
and IoU threshold $\tau$ used in LSA.  
Ours performs robustly under a wide range of settings; the chosen ($L{=}20$, $\tau{=}0.3$) strikes a good balance.

\vspace{-1mm}
\section{Conclusion}
\label{sec:conclusion}
We presented \shortname, a training-free streaming reconstruction framework that converts an offline 4D reconstruction model into a streaming system. 
By introducing layer-wise scale alignment, we address the key challenge of inconsistent depth scaling across temporal windows, enabling stable alignment and long-range geometric consistency.
Extensive experiments demonstrate that \shortname~achieves state-of-the-art camera pose estimation accuracy, reconstruction quality, speed and memory budget.
We believe this work provides a new angle towards bridging offline and streaming reconstruction. 
We hope it will inspire future research on integrating classical geometric principles with modern neural architectures for large-scale, continuous 3D perception.

\clearpage

\section{Acknowledgment}
\label{sec:acknowledge}
Tianye Ding and Huaizu Jiang were partially supported by the National Science Foundation under Award IIS-2310254.
Yiming Xie was supported by the Apple Scholars in AI/ML PhD fellowship.
Pedro Miraldo and Moitreya Chatterjee were supported exclusively by Mitsubishi Electric Research Laboratories.
{
    \small
    \bibliographystyle{ieeenat_fullname}
    \bibliography{main}
}

\clearpage
\setcounter{page}{1}
\maketitlesupplementary

\section{More Details about Submap Registration in \texorpdfstring{$\mathrm{Sim}(3)$}{Sim(3)} Space}
\label{sec:registration}

3D reconstruction in each window $\calW_i$ yields a local submap 
$\mathcal{S}_i=\{\bTti \bPti\}_{t\in \calW_i}$ 
in the window's own coordinate system.
We then estimate a similarity transform 
$(s_i^w,\bRiw,\btiw)\in\mathrm{Sim}(3)$ between $\calS_i$ to $\calG_{i-1}$, which are defined in the world coordinate system (in our case, the first temporal window's coordinate system, based on the estimated point maps of the overlapping region.
The induced camera pose in the world space for a frame $\bI_{t\in \calW_i}$ is
$\bT_{t}^{w} = (\bRiw \bRti |\ s_i^w \bRiw \btti + \btiw )$
The global map $\calG_i$ is then updated progressively as $\calG_i = \calG_{i-1} \cup \{\bT_{t}^{w} \bPti\}$, where $\calG_0=\emptyset$ in the initialization.


To estimate the $\mathrm{Sim}(3)$ transform, 
we first estimate the global scale factor $\siw$ via a robust IRLS (Iteratively Reweighted Least Squares) optimization, enforcing a shared metric across two adjacent windows.
Rotation and translation $(\bRiw,\btiw)$ are then optimized via the Kabsch algorithm ~\cite{Kabsch} under that metric using the \emph{scaled} camera anchors based on the estimated $\siw$.


\vspace{0pt}
\noindent\textbf{Scale estimation via IRLS based on point correspondences.}
We estimate the per-window scale $\siw$ from point-wise correspondences in the intersection of two consecutive windows.
Specially, 
for overlapping frames that share the same timestamp $t$ in $\calW_{i-1}$ and $\calW_i$, 
we extract 3D points for every pixel $x$ in the intersection of the two windows
$
\bp(x) = \bPtiminusone (x),
\bq(x) = \bPti (x),
$
and their associated confidences
$c_p(x) = \bCtiminusone(x)$, $c_q(x) = \bCti(x)$.
The set of \emph{mutually confident correspondences} is then defined as:\footnote{To avoid notation clutter, we omit the variable $x$ from now on.}
\begin{equation}
\mathcal{C} = 
\{\, (\bp, \bq) \mid 
c_p > g(\bCtiminusone),\,
c_q > g(\bCti) \,\},
\label{eq:confidence}
\end{equation}
where $g$ denotes the median function.
Each pair $(\bp,\bq)\!\in\!\mathcal{C}$ represents the same 3D point in two coordinates of submaps, with both predictions considered reliable.
We estimate the optimal scale $\siw$ by solving the Huber-robust objective:
\begin{equation}
\siw = \arg\min_{s>0}\;
\sum_{(\bp,\bq)\in\mathcal{C}}
\rho~\!\bigl(\|\,s\,\bp - \bq\,\|_2\bigr),
\label{eq:scale_irls}
\end{equation}
where $\rho(\cdot)$ is the Huber loss with parameter $\delta$.


\vspace{0pt}
\noindent\textbf{Rotation and translation based on scaled camera anchors.}
After estimating the global scale $\siw$ from confident point correspondences, we scale the submap $\calS_i$ first and then estimate the rigid transformation.
We define canonical camera axes in each camera's coordinate system as the \emph{up} $\bu=(0,1,0)$ and \emph{view} $\bv=(0,0,-1)$.
Let $\mathcal{O}_i \!=\! \calW_{i-1} \cap \calW_i$ be the set of overlapping timestamps. Using the camera center $\btti$ and normalized axes $(\bv_t,\bu_t)$, 
we form two scaled camera anchor sets $\{\vx_t\}_{t\in\calO_i}$ and $\{\vy_t\}_{t\in\calO_i}$, where:
\begin{align}
\vx_t &= ( 
\siw\,\bt_t^{(i)},\;
\siw\,\bt_t^{(i)}+\bv_t^{(i)},\;
\siw\,\bt_t^{(i)}+\bu_t^{(i)}),\\
\vy_t &= ( 
\bt_t^{(i-1)},\;
\bt_t^{(i-1)}+\bv_t^{(i-1)},\;
\bt_t^{(i-1)}+\bu_t^{(i-1)}).
\label{eq:anchor_sets}
\end{align}
We then estimate the window-level rigid transform $(\bRiw,\btiw)$ by minimizing
the alignment error between the two anchor sets via the Kabsch algorithm~\cite{Kabsch}:
\begin{equation}
\bRiw,\,\btiw
=\arg\min_{\bR\in SO(3),\,\bt\in\mathbb{R}^3}
\sum_{t\in\calO_i}\bigl\|\,\bR\, \vx_t + \bt - \vy_t \bigr\|_2^2.
\label{eq:kabsch_objective}
\end{equation}

\noindent\textbf{Differences from existing approaches.} 
Although VGGT-Long~\cite{deng2025vggtlongchunkitloop} also adopts a sliding-window strategy for streaming inputs, our method differs in how the registration $\mathrm{Sim}(3)$ is estimated from overlapping windows. VGGT-Long applies IRLS to jointly optimize a closed-form scale $s$ together with $\bR$ and $\bt$ computed via the Kabsch algorithm. In contrast, we first estimate the scale using point-cloud correspondences within matched camera coordinate systems, and then estimate $\bR$ and $\bt$ using the scaled inputs. This two-stage procedure yields more stable and robust registration.

Furthermore, our $\mathrm{SE}(3)$ registration is obtained from minimal camera anchors derived directly from camera poses. These anchors avoid the artifacts introduced by point-map predictions and additionally preserve trajectory consistency, particularly in small-scale scenes.

We conduct ablation studies on these registration strategies in \cref{sec:ablation_submap_reg}.

\begin{table*}[t]
\centering
\setlength{\tabcolsep}{4pt}
\footnotesize
\vspace{-3mm}
\caption{Outdoor, Long-term Point Map Estimation on \textbf{Waymo}~\cite{waymo}. We report Accuracy (Acc, lower is better), Completeness (Comp, lower is better) and Chamfer Distance (Chamfer, lower is better). We show metrics for each segment ID; \emph{Avg.} is the mean across segments. 
}
\vspace{-2mm}
\resizebox{\textwidth}{!}{
\begin{tabular}{l l |c| c *{9}{c}}
\toprule
\textbf{Segment ID} & \textbf{Metric} & \textbf{Avg.}
& \textbf{163453191} & \textbf{183829460} & \textbf{315615587} & \textbf{346181117}
& \textbf{371159869} & \textbf{405841035} & \textbf{460471311} & \textbf{520018670} & \textbf{610454533} \\
\midrule

\multirow{3}{*}{VGGT\textendash Long~\cite{deng2025vggtlongchunkitloop}}
& Acc $\downarrow$     & 0.508 & 0.453 & 0.096 & 0.629 & 1.441 & 0.457 & 0.379 & 0.510 & 0.481 & 0.129 \\
&  Comp $\downarrow$ & 0.456 & 0.412 & 0.101 & 0.552 & 1.341 & 0.365 & 0.344 & 0.496 & 0.386 & 0.102 \\
& Chamfer $\downarrow$      & 0.482 & 0.432 & 0.098 & 0.591 & 1.391 & 0.411 & 0.361 & 0.503 & 0.434 & 0.115 \\
\midrule
\multirow{3}{*}{$\pi^3$-Long}
& Acc $\downarrow$     & 1.043 & 0.912 & 0.160 & 1.209 & 0.837 & 1.728 & 0.228 & 0.545 & 3.101 & 0.668 \\
&  Comp $\downarrow$ & 0.745 & 0.738 & 0.145 & 0.502 & 0.362 & 1.085 & 0.155 & 0.208 & 3.149 & 0.356 \\
& Chamfer $\downarrow$      & 0.894 & 0.825 & 0.153 & 0.856 & 0.600 & 1.406 & 0.192 & 0.376 & 3.125 & 0.512 \\
\midrule
\multirow{3}{*}{\textbf{Ours ($\pi^3$)}}
& Acc $\downarrow$     & 0.560 & 0.422 & 0.176 & 1.151 & 0.651 & 0.896 & 0.127 & 0.541 & 0.247 & 0.832 \\
&  Comp $\downarrow$ & 0.266 & 0.315 & 0.158 & 0.194 & 0.223 & 0.459 & 0.106 & 0.326 & 0.232 & 0.385 \\
& Chamfer $\downarrow$      & 0.413 & 0.368 & 0.167 & 0.673 & 0.437 & 0.677 & 0.116 & 0.434 & 0.240 & 0.608 \\
\bottomrule
\end{tabular}
}
\vspace{0mm}
\label{tab:mvrecon_segments}
\end{table*}

\begin{figure*}[!ht]
    \centering
\includegraphics[width=0.9\linewidth]{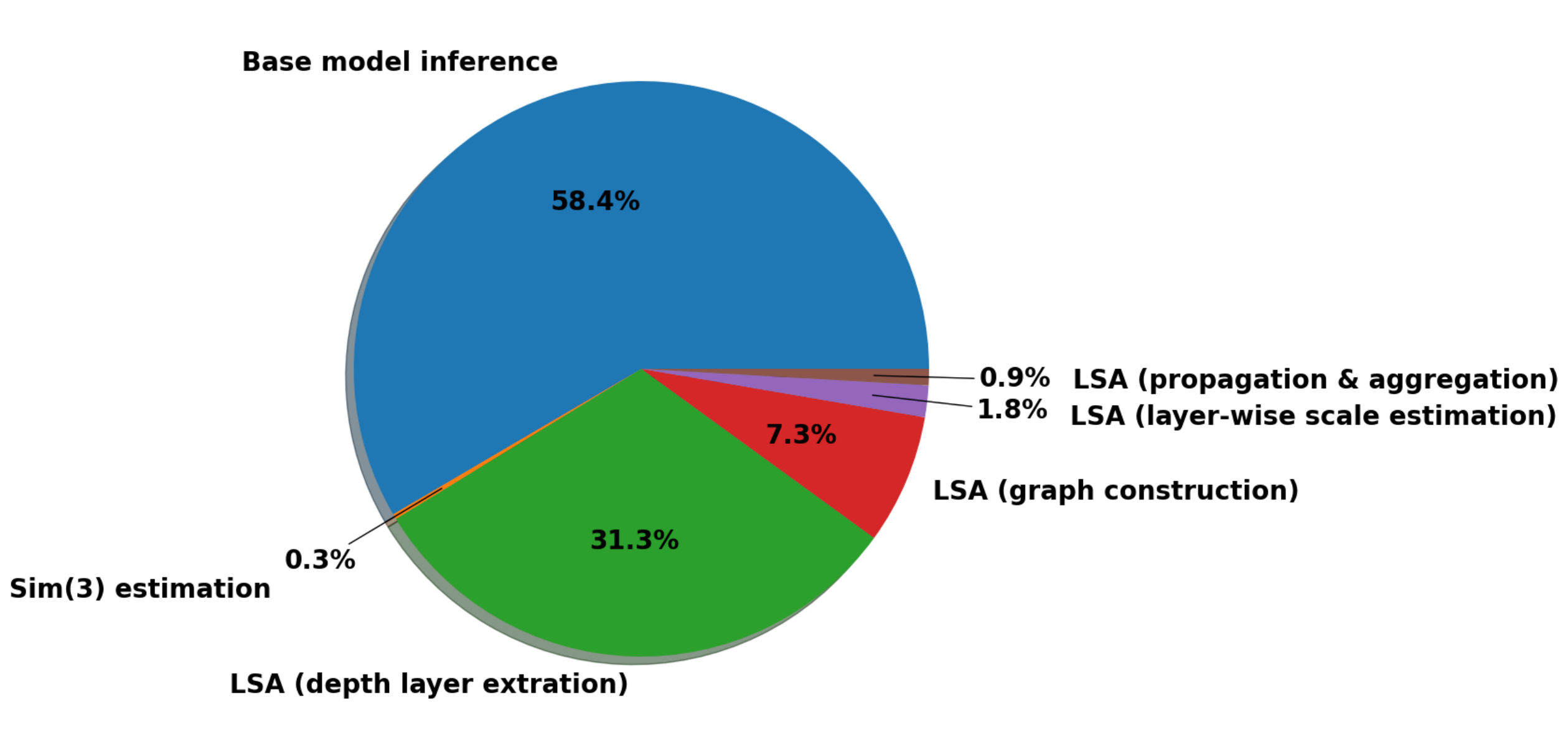}
    \caption{
        Runtime analysis of each module within the pipeline.
    }
\label{fig:runtime_pie}
\end{figure*}

\section{Implementation Details}

We instantiate \shortname\ using either VGGT~\cite{wang2025vggt} or $\pi^3$~\cite{wang2025pi} as the offline 3D reconstruction backbone. 
For video depth estimation, small-scale camera pose estimation, and indoor multi-view point map estimation, we evaluate both variants. 
For large-scale camera pose estimation on KITTI~Odometry~\cite{kitti} and outdoor point map estimation on Waymo~\cite{waymo}, we use $\pi^3$ as the backbone for its stronger geometric prior. 
On the kilometer-scale KITTI~Odometry, we additionally incorporate loop closure following the VGGT-Long~\cite{deng2025vggtlongchunkitloop} configuration for fairness.

We use multi-threading to run model inference for each window and registration of adjacent window pairs with LSA refinement concurrently.
At the beginning of depth graph construction, we try to assign each frame to separate available threading to achieve maximum parallelism.

\begin{table*}[t]
\centering
\setlength{\tabcolsep}{6pt}
\caption{Ablation Studies for Submap Registration~(\cref{sec:registration}). 
\emph{w$/$o IRLS} denotes estimating scale via closed-form solution instead of IRLS.
\emph{w$/$o Anchor} denotes estimating rigid transformation on scaled point maps instead of scaled camera anchors.
}
\resizebox{.7\textwidth}{!}{
\begin{tabular}{lcc|cc|ccc}
\toprule
& \multicolumn{2}{c|}{\textbf{Sintel}} & \multicolumn{2}{c|}{\textbf{Bonn}} & \multicolumn{3}{c}{\textbf{Sintel}} \\
\cmidrule(lr){2-3} \cmidrule(lr){4-5} \cmidrule(lr){6-8}
 & Abs Rel $\downarrow$ & $\delta{<}1.25$ $\uparrow$ & Abs Rel $\downarrow$ & $\delta{<}1.25$ $\uparrow$ &
ATE $\downarrow$ & RPE$_{\text{trans}}$ $\downarrow$ & RPE$_{\text{rot}}$ $\downarrow$ \\
\midrule 
\textbf{Ours} & \textbf{0.247} & \textbf{68.8} & \textbf{0.048} & \textbf{97.4} & \textbf{0.061} & \textbf{0.028} & \textbf{0.249} \\
w$/$o IRLS & 0.328 & 51.4 & 0.123 & 85.6 & 0.107 & 0.035 & 0.249 \\
w$/$o Anchor & 0.247 & 68.8 & 0.048 & 97.4 & 0.081 & 0.039 & 0.742 \\
\bottomrule
\end{tabular}}
\label{tab:ablation_registration}
\end{table*}
\begin{figure}[t]
    \centering
\includegraphics[trim={0 0pt 0 0pt}, clip, width=1.0\linewidth]{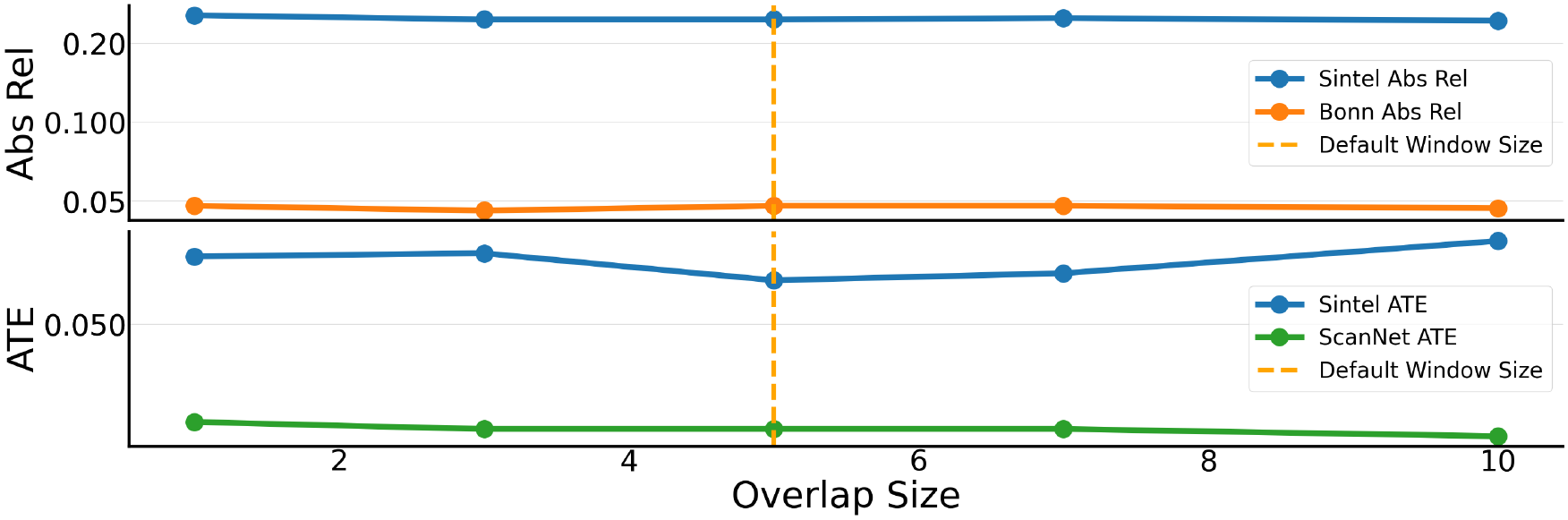}
    \vspace{-8mm}
    \caption{
        \textbf{Ablation on overlap size $O$.} In default, we use $O=5$.
    }
    \vspace{-5mm}
\label{fig:ablation_overlap}
\end{figure}

\section{Submap Registration (\cref{sec:registration}).}
\label{sec:ablation_submap_reg}
\begin{figure*}[t]
\centering
\begin{minipage}[t]{0.33\textwidth}
\centering
\includegraphics[width=\linewidth]{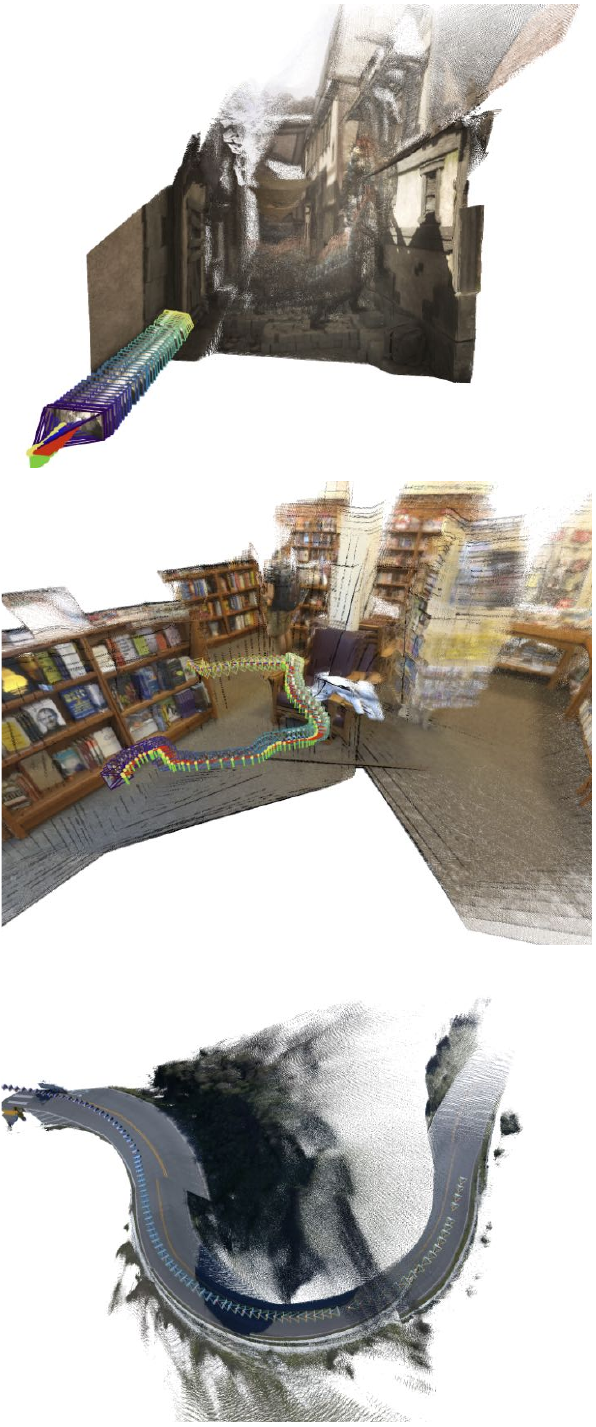}
w/o IRLS
\end{minipage}
\begin{minipage}[t]{0.33\textwidth}
\centering
\includegraphics[width=\linewidth]{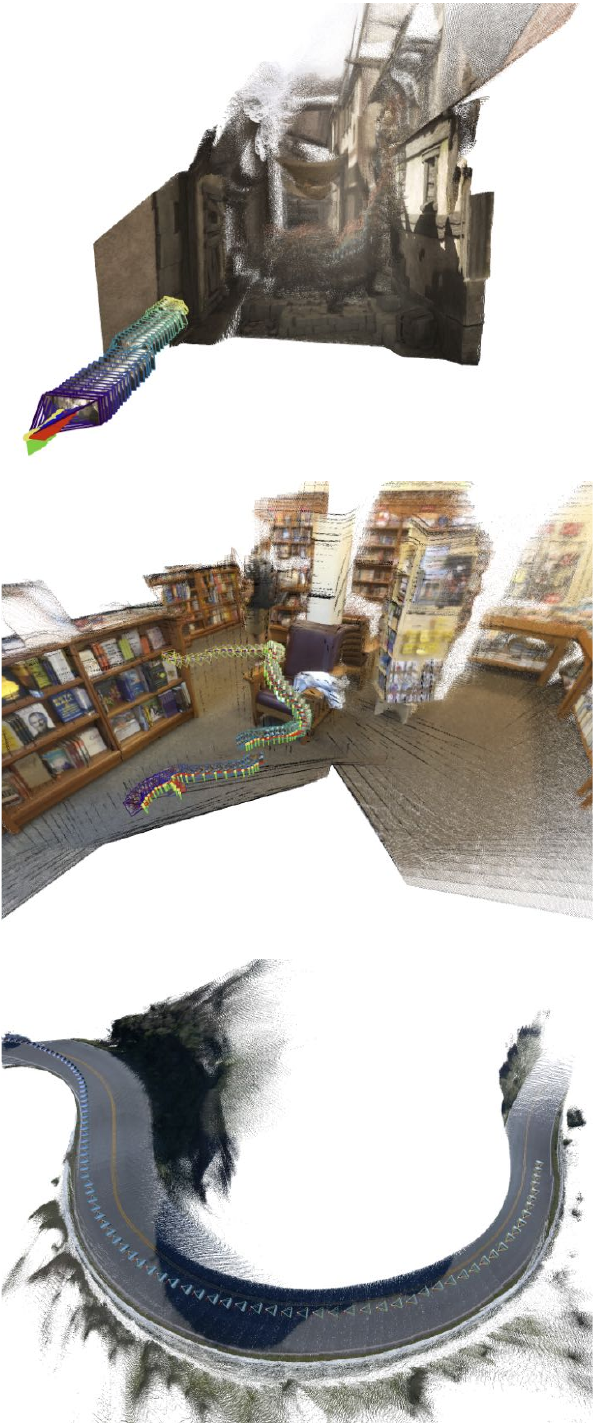}
w/o Anchor
\end{minipage}
\begin{minipage}[t]{0.33\textwidth}
\centering
\includegraphics[width=\linewidth]{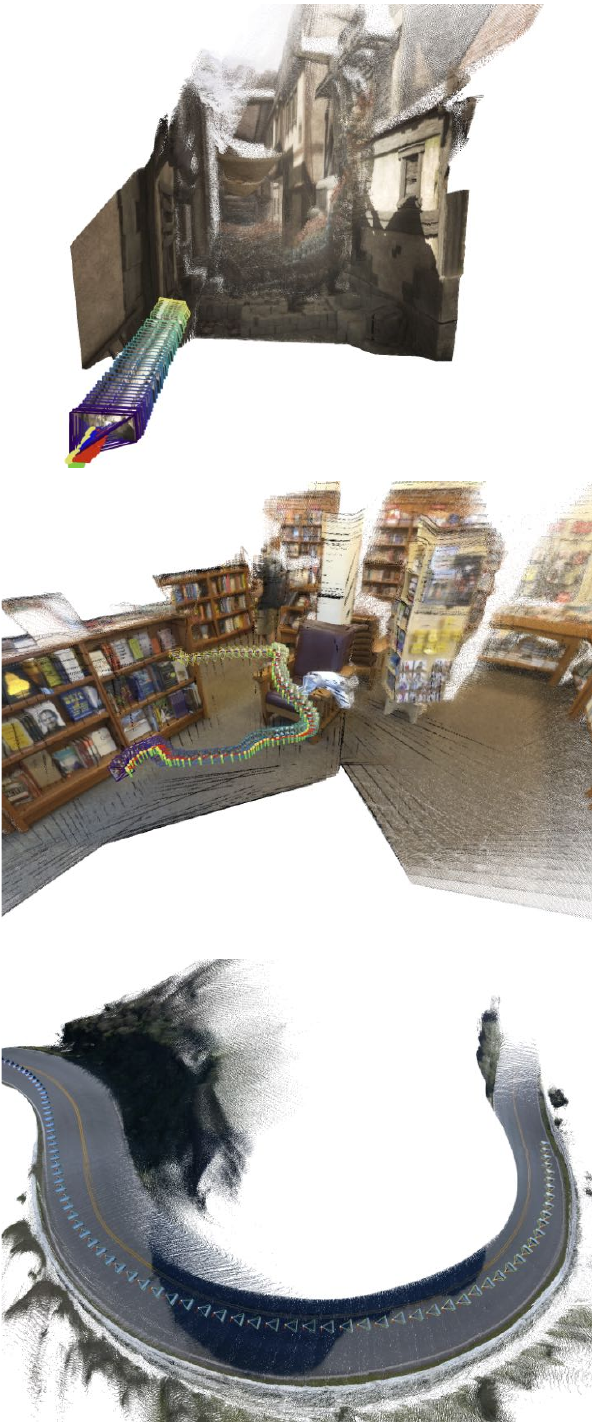}
\textbf{Ours}~($\pi^3$)
\end{minipage}
\caption{Ablation Studies for Submap Registration~(\cref{sec:registration}).
\emph{w$/$o IRLS} denotes estimating scale via closed-form solution instead of IRLS.
\emph{w$/$o Anchor} denotes estimating rigid transformation on scaled point maps instead of scaled camera anchors.
}
\label{fig:ablation_submap_reg}
\end{figure*}

\cref{fig:ablation_overlap} examines the effect of selecting different overlap sizes $O$, our method is robust to a wide range of settings, the default choice ($O=5$) maintains a good balance between accuracy and inference complexity.
\cref{fig:ablation_submap_reg} and \cref{tab:ablation_registration} compares alternative strategies for estimating the $\mathrm{SE}(3)$ transform between submaps.  
Replacing IRLS with a closed-form solver degrades both depth and pose accuracy, confirming the importance of robust scale estimation in this stage.
Replacing scaled camera anchors with scaled point maps produces similar depth metrics but noticeably weaker camera trajectories.

\section{Time Analysis for the Registration Module}

We also provide a detailed runtime analysis of each module in our framework, as shown in Fig.~\ref{fig:runtime_pie}. 
Using a window size of $20$ with an overlap of $5$, the measured runtimes are as follows:
$\pi^3$ single inference pass: $1.344\,\mathrm{s}$; 
$\mathrm{Sim}(3)$ estimation: $0.007\,\mathrm{s}$; 
depth-layer extraction: $0.719\,\mathrm{s}$; 
graph construction: $0.168\,\mathrm{s}$; 
scale initialization: $0.041\,\mathrm{s}$; 
and propagation \& aggregation: $0.021\,\mathrm{s}$.

\section{Evaluation Details of Efficiency Benchmark}
We report FPS and peak memory usage on the Sintel~\cite{sintel} benchmark for all methods on an A6000 GPU. The image resolution for DUSt3R-based~\cite{dust3r} methods is $512 \times 288$ except Spann3R, which only supports $224 \times 224$, and VGGT-based~\cite{wang2025vggt} methods are $518 \times 294$.

\section{Outdoor Multi-view Point Map Estimation}
Tab.~\ref{tab:mvrecon_segments} reports long-term multi-view point map estimation results. 
\shortname\ using $\pi^3$ as backbone achieves the best overall performance among training-free methods, substantially outperforming both VGGT-Long~\cite{deng2025vggtlongchunkitloop} and $\pi^3$-Long in Comp and Chamfer while maintaining comparable Acc. 

For outdoor setting, we use the Waymo Open Dataset~\cite{waymo} on urban driving segments 
and report Acc, Comp, and Chamfer distance, 
following
~\cite{deng2025vggtlongchunkitloop} (results in the supplementary material). 
To mitigate artifacts from sky and far-background regions, we uniformly filter out the lowest-confidence 40\% of predicted points for \emph{all} methods; these results serve as a comparative reference rather than a strict head-to-head benchmark.

\begin{figure*}[t]
\centering
\begin{minipage}[t]{0.24\textwidth}
\centering
\includegraphics[width=\linewidth]{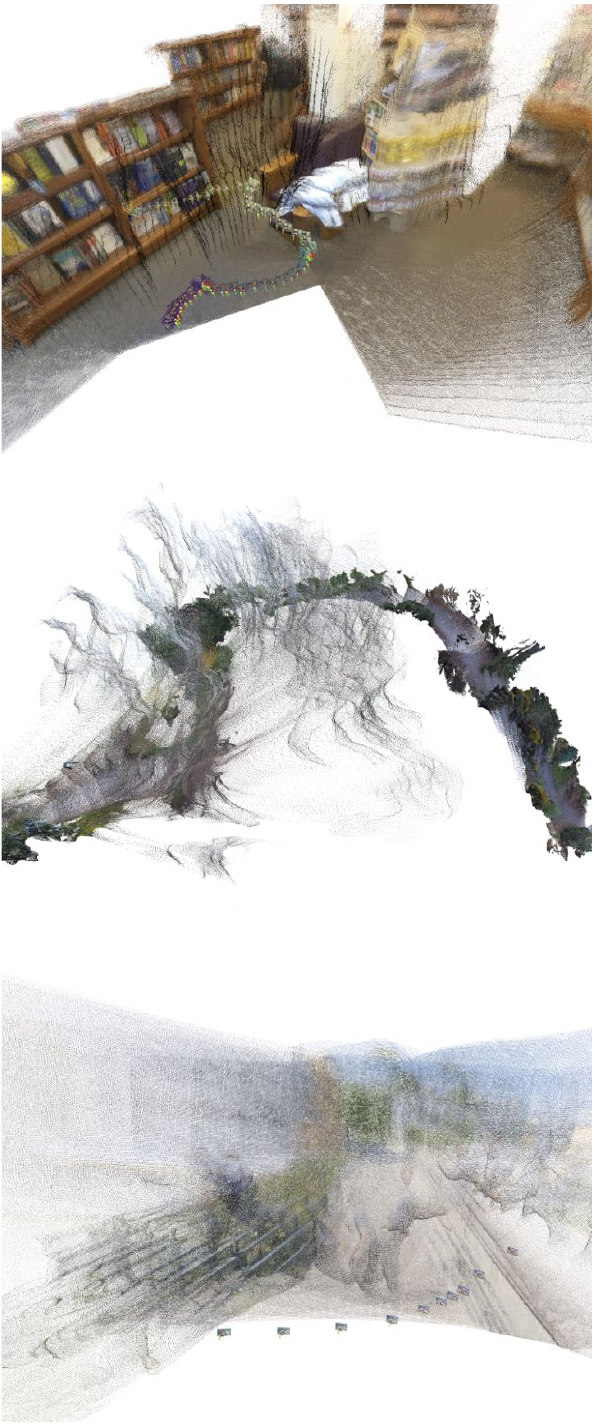}
CUT3R~\cite{wang2025cut3r}
\end{minipage}
\begin{minipage}[t]{0.24\textwidth}
\centering
\includegraphics[width=\linewidth]{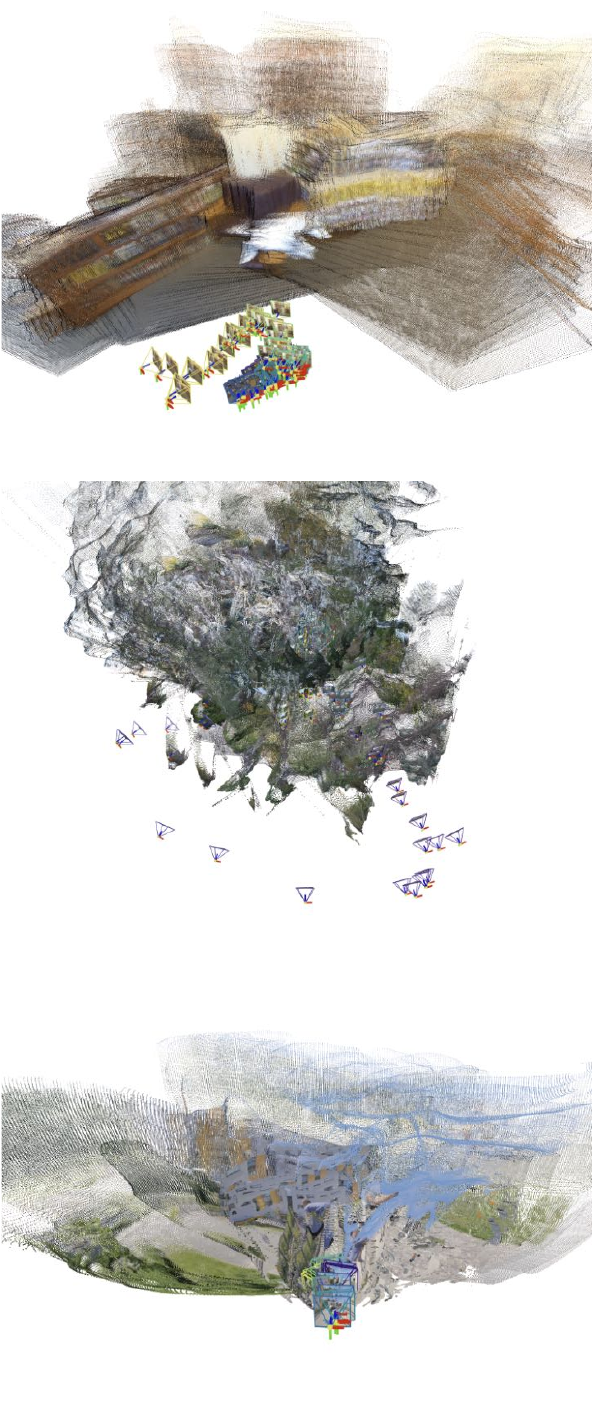}
StreamVGGT~\cite{zhuo2025streaming}
\end{minipage}
\begin{minipage}[t]{0.24\textwidth}
\centering
\includegraphics[width=\linewidth]{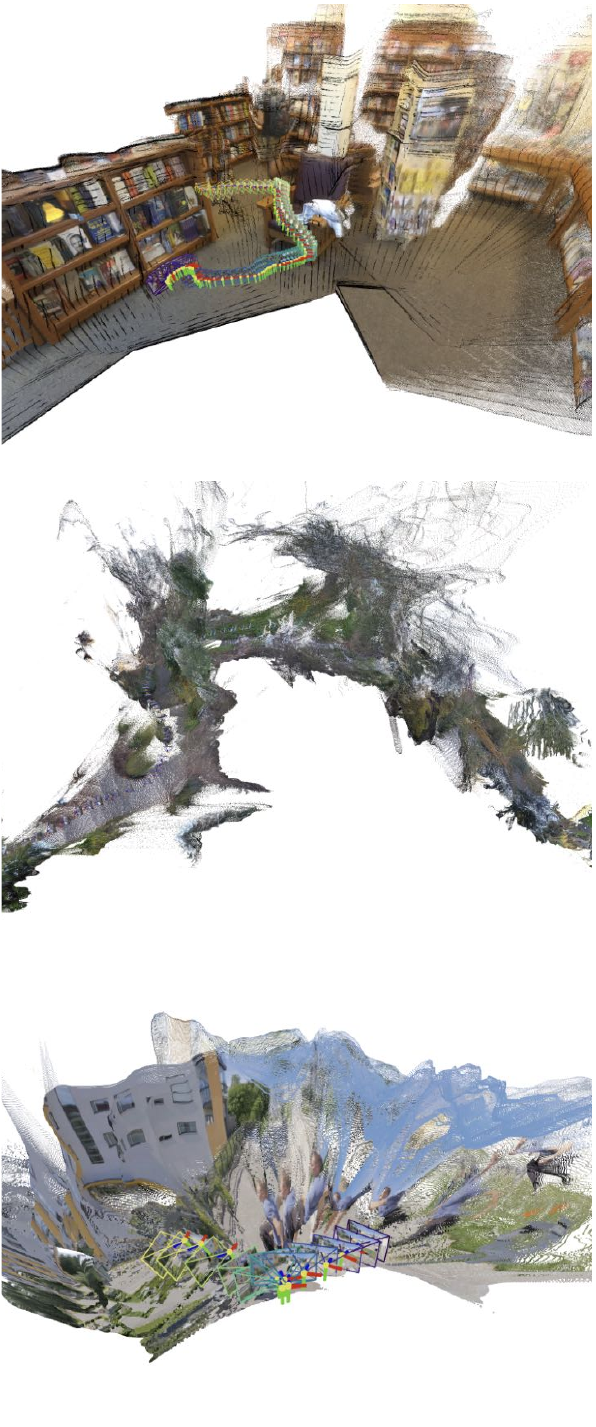}
VGGT-SLAM~\cite{vggtslam}
\end{minipage}
\begin{minipage}[t]{0.24\textwidth}
\centering
\includegraphics[width=\linewidth]{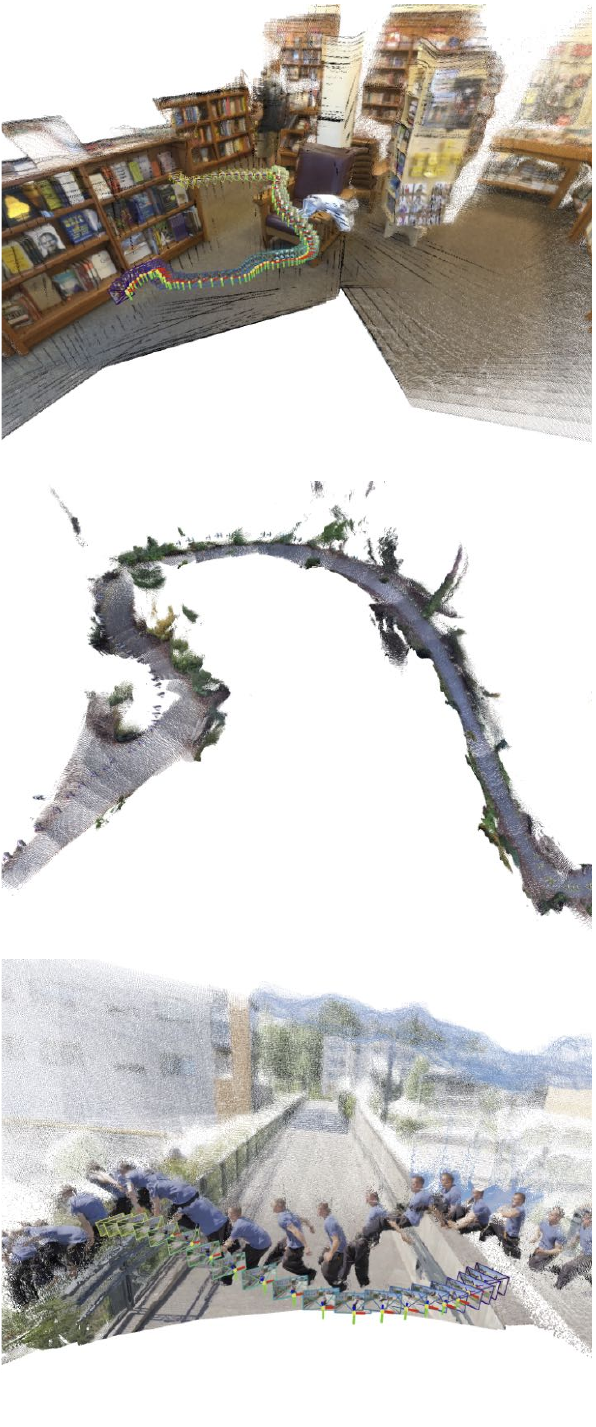}
\textbf{Ours}~($\pi^3$)
\end{minipage}
\caption{Qualitative comparison on different sequences.}
\label{fig:supp_qualitative}
\end{figure*}

\begin{figure*}[t]
    \centering
    \includegraphics[width=1.0\linewidth, trim={0 0 15pt 0}, clip]{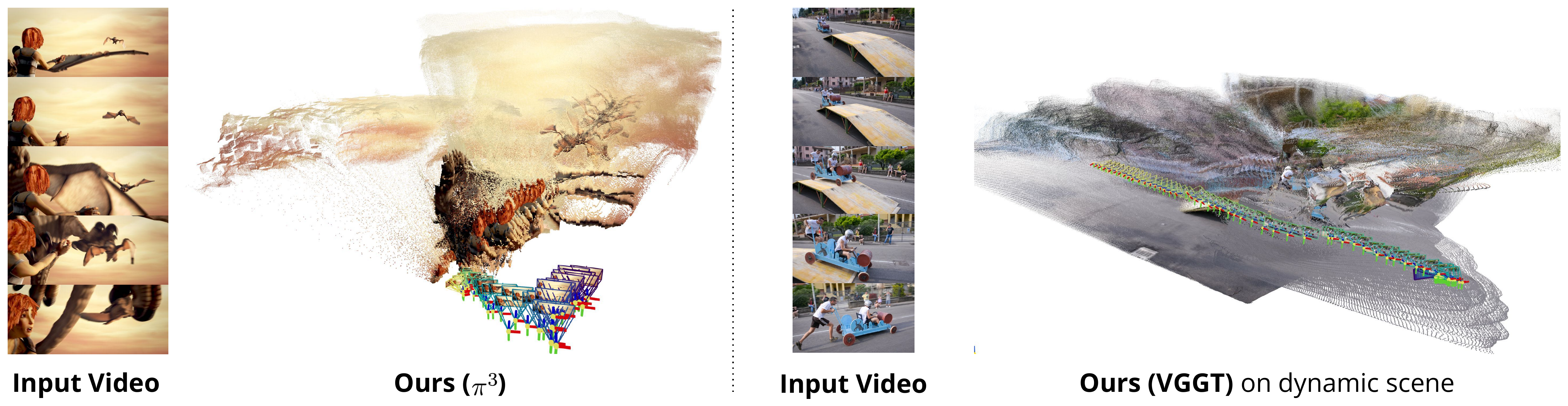}
    \caption{
        Failure Cases. 
    }
    \vspace{-4mm}
\label{fig:failure}
\end{figure*}
\section{Future Directions}
Although our method demonstrates strong performance, it has room for improvements. We show some failure cases in \cref{fig:failure}, and list three directions that interest us most:
\begin{itemize}
    \item \textbf{Different hyperparameters for indoor and outdoor scenes.}
    Our framework requires empirical hyperparameter tuning for diverse environments (e.g., window size, overlap ratio, and depth-layer confidence thresholds). While this manual tuning improves stability for each domain, it reduces the generality of our method and makes adaptive adjustment when transferred to new settings an interesting direction to explore.

    \item \textbf{Performance bounded by backbone reconstructors.}
    Because our system is built on top of offline 3D reconstructors, its performance is heavily dependent on the backbone submap prediction quality. For example, when using VGGT as backbone, our method inherits VGGT’s inability to handle dynamic or non-rigid scenes. As VGGT struggles to maintain reliable geometry and camera pose estimates in the presence of moving objects, our method also fails under such conditions. This dependency limits applicability to fully static or quasi-static scenes. We look forward to seeing how advancement on offline 3D reconstructors can boost our method as well.

    \item \textbf{LSA Sensitivity to complex scenes and object occlusions.}
    Our current depth layer graph construction used in LSA is based purely on predicted frame-wise depth maps, without persistent contexts of distinct objects, which results in multiple objects being assigned to the same depth layer within complex scenes, also unable to recover layer correspondences if an object is occluded entirely and later re-enters the scene.
    We seek to improve the robustness of our design and the temporal trackability of different objects by incorporating appearance information from raw images.
\end{itemize}

\end{document}